\documentclass[letterpaper, 10 pt, conference]{ieeeconf}
\usepackage{cite}
\usepackage{amsmath,amssymb,amsfonts} 
\usepackage{graphicx}
\usepackage{textcomp} 
\usepackage{multirow}
\usepackage{gensymb}
\usepackage{xcolor} 
\usepackage[ruled,vlined]{algorithm2e}
\usepackage{gensymb}
\usepackage{textcomp} 
\usepackage{xcolor} 
\usepackage{color}
\usepackage{multirow}
\usepackage{graphicx}
\usepackage{subfigure}
\usepackage{bm}
\usepackage[colorlinks,linkcolor=blue]{hyperref}

\IEEEoverridecommandlockouts                              

\makeatletter

\makeatletter
\renewcommand{\maketag@@@}[1]{\hbox{\m@th\normalsize\normalfont#1}}%

\setbox0\hbox{$\xdef\scriptratio{\strip@pt\dimexpr
    \numexpr(\sf@size*65536)/\f@size sp}$}

\newcommand{\scriptveryshortarrow}[1][3pt]{{%
    \vcenter{\hbox{\rule[\scriptratio\dimexpr-.2pt\relax]
               {\scriptratio\dimexpr#1\relax}{\scriptratio\dimexpr.4pt\relax}}}%
   \mkern-4mu\hbox{\let\f@size\sf@size\usefont{U}{lasy}{m}{n}\symbol{41}}}}

\makeatother

\title{\LARGE \bf
OpenCalib: A Multi-sensor Calibration Toolbox for Autonomous Driving
}

\author{Guohang Yan$^{1}$, Liu Zhuochun$^{1}$, Chengjie Wang$^{1}$, Chunlei Shi$^{1}$, Pengjin Wei$^{1}$, Xinyu Cai$^{1}$ \\Tao Ma$^{1}$, Zhizheng Liu$^{1}$, Zebin Zhong$^{2}$, Yuqian Liu$^{2}$, Ming Zhao$^{2}$, Zheng Ma$^{2}$ and Yikang Li$^{\dagger 1 2}$ 
\thanks{$^{\dagger}$ Corresponding author.}
\thanks{$^{1}$ Guohang Yan, Liu Zhuochun, Chengjie Wang, Chunlei Shi, Pengjin Wei, Xinyu Cai, Tao Ma, Zhizheng Liu and Yikang Li are with Autonomous Driving Group, Shanghai AI Laboratory, China. {\tt\small \{yanguohang, liuzhuochun, wangchengjie, shichunlei, weipengjin, caixinyu, matao, liuzhizheng, liyikang\}@pjlab.org.cn}}
\thanks{$^{2}$ Zebin Zhong, Yuqian Liu, Ming Zhao, Zheng Ma and Yikang Li are with Autonomous Driving Group, SenseTime, China. {\tt\small \{zhongzebin, liuyuqian, zhaoming, mazheng, liyikang\}@senseauto.com}}
}

\begin{document}
 
\maketitle

\begin{abstract}
Accurate sensor calibration is a prerequisite for multi-sensor perception and localization systems for autonomous vehicles. The intrinsic parameter calibration of the sensor is to obtain the mapping relationship inside the sensor, and the extrinsic parameter calibration is to transform two or more sensors into a unified spatial coordinate system.
Most sensors need to be calibrated after installation to ensure the accuracy of sensor measurements. To this end, we present OpenCalib, a calibration toolbox that contains a rich set of various sensor calibration methods. OpenCalib covers manual calibration tools, automatic calibration tools, factory calibration tools, and online calibration tools for different application scenarios. At the same time, to evaluate the calibration accuracy and subsequently improve the accuracy of the calibration algorithm, we released a corresponding benchmark dataset. This paper introduces various features and calibration methods of this toolbox. To our knowledge, this is the first open-sourced calibration codebase containing the full set of autonomous-driving-related calibration approaches in this area. 
We wish that the toolbox could be helpful to autonomous driving researchers. We have open-sourced our code on GitHub to benefit the community. Code is available at \href{https://github.com/PJLab-ADG/SensorsCalibration}{https://github.com/PJLab-ADG/SensorsCalibration}.
 \end{abstract}

\section{INTRODUCTION}
Autonomous driving has recently become one of the most popular technologies in both industry and academia. As a complicated system, it requires numerous modules to collaboratively work together.
Since different sensors have their advantages and shortcomings, fusing multiple heterogeneous sensors becomes the key to robust and accurate perception and localization ability. 
IMU (Inertial measurement unit), GNSS (Global Navigation Satellite System), LiDAR (Light Detection and Ranging), Camera, millimeter-wave Radar, and wheel speedometers are most-commonly used sensors for autonomous driving systems. 

IMU is the inertial navigation sensor that can provide relative displacement and heading angle changes with high confidence in a short time. As we all know, GNSS can provide absolute positioning for vehicles with meter-level accuracy. However, the quality of GNSS signals can not be guaranteed at all times. Therefore, in the field of autonomous driving, the output of GNSS is generally fused with the IMU and the car's sensors (such as wheel speedometers, steering wheel angle sensors, etc.).

The camera has a relatively wide range of applications and a relatively low cost among the many mainstream sensors. The camera has a solid ability to extract detailed information about the environment. In addition to color, it can also provide texture and contrast data—reliable identification of road markings or traffic signs, accurate detection and identification of stationary and moving objects. So, no matter which sensor scheme is used, the camera is generally used. The technology of processing images by relying on powerful processors, algorithms and neural networks are relatively mature. Despite so many advantages, the camera is unable to "take the lead" in the autonomous driving perception system. This is not only because it requires a good lighting environment. The camera's reliability is limited in harsh environmental conditions such as snow or fog and the dark. It is also because the data or image it obtains is two-dimensional, without direct depth value information. The depth of information obtained by the algorithm is not accurate enough. 

LiDAR has outstanding advantages such as high accuracy, long-ranging, good real-time performance, and a rich collection of information. At the same time, it has better environmental adaptability and is not affected by light. However, the cost of LiDAR is relatively high and generally expensive. Another, it lacks color information and is not accurate enough to detect objects that can produce reflection or transparency. The collected data requires extremely high computing power, and the scanning speed is relatively slow. Millimeter-wave radar is also a class of environmental sensors with relatively mature technology. Due to its mass production, Radar is relatively inexpensive, has high detection accuracy for surrounding vehicles, and is sensitive to certain materials. At the same time, it responds quickly, is easy to operate, and can adapt to bad weather. However, the resolution of Radar is relatively low. It cannot judge the size of the recognized object and perceive pedestrians. It cannot accurately model all surrounding obstacles. Due to the advantages and disadvantages of a single sensor, the purpose of the multi-sensor fusion system is to improve information redundancy and information complementarity so that the safety of autonomous driving can be fully guaranteed.

The previous paragraph introduced the advantages and disadvantages of each sensor in autonomous driving and the necessity of multi-sensor fusion. However, if you want a good fusion result, these different sensors need to be calibrated accurately. There are multiple sensors installed on a car, and the coordinate relationship between them needs to be determined by sensors calibration. Therefore, sensors calibration is the basic requirement of automatic driving. Sensors calibration can be divided into two parts: intrinsic parameter calibration and extrinsic parameter calibration. The intrinsic parameter determines the internal mapping relationship of the sensor. For example, the intrinsic camera parameters are calibrated with the focal length and lens distortion. The IMU intrinsic parameters are calibrated with the zero bias of gyroscope and accelerometer, scale factor and installation error. The LiDAR intrinsic parameters are the conversion relationship between the internal laser transmitter coordinates and the LiDAR coordinate device. The extrinsic parameter determines the conversion relationship between the sensor and an external coordinate system, including 6 degrees of freedom parameters for rotation and translation.

In the research and development of autonomous driving, the calibration of extrinsic parameters between sensors is common. The biggest problem in calibrating between different sensors is how to measure the best, because the types of data obtained are different. Therefore, the objective function for minimizing the calibration error will be different due to different sensor pairings. The extrinsic calibration methods can generally be classified into target-less and target-based. The former is carried out in a natural environment with few constraints and does not require a special target; the latter requires a special control field and has a ground truth target.  

At present, there are some open-source projects related to sensor calibration, such as \href{https://github.com/ethz-asl/kalibr}{Kalibr}, Autoware \cite{kato2018autoware}, etc, and there are more open-source projects of a specific calibration type, such as \href{https://github.com/ethz-asl/lidar_align}{lidar\_align} for lidar and IMU calibration. OpenCV \cite{opencv_library} also provide some calibration tools. However, there is currently no complete calibration toolbox for different application scenarios of autonomous driving. So we integrated various calibration tools for autonomous driving into a calibration toolbox based on previous calibration research and project experience.

The contributions of this work is listed as follows: 
\begin{enumerate}
\item We propose OpenCalib, a multi-sensor calibration toolbox for autonomous driving. The toolbox can be used for different calibration scenarios, including manual calibration tools, automatic calibration tools, factory calibration tools and online calibration tools.
\item We propose many novel calibration algorithms in the toolbox, such as various automatic calibration methods based on road scenes. For the factory calibration in the toolbox, we propose more robust recognition algorithms for multiple calibration board types and the calibration board recognition program removes the OpenCV \cite{opencv_library} library dependency.
\item To benchmark the calibration performance, we introduce a synthesized dataset based on Carla \cite{Dosovitskiy17}, where we can get the ground truth for the calibration results. In the future, we will further open-source the calibration benchmark dataset. 
\item We open-source the toolbox code on GitHub to benefit the community. The open-source code is the v0.1 version, and we will continue introducing more state-of-the-art calibration algorithms in future versions. 
\end{enumerate}


\section{RELATED WORK}
Researchers have proposed many approaches to address the calibration problem of multi-model sensor calibration.
According to the observed scene, these calibration methods are broadly classified into target-based and targetless-based methods (with and without artificial targets). It is further divided into manual and automatic methods according to the level of human interaction. In addition, the researchers also proposed some online calibration methods, mainly motion-based methods based on motion estimation and learning-based methods based on deep learning. 

\subsection{Target-based Method}
Target-based calibration methods are widely used in the sensor calibration process. Target-based methods often require manual calibration targets, such as chessboards, polygonal board, which both sensor modalities can easily detect. 
Furthermore, target-based methods can use prior knowledge about the target, enhancing calibration results. The target methods are more precise than target-less. 
The intrinsic calibration of the sensor is usually by the target method. The commonly used camera intrinsic calibration method is Zhang's method \cite{zhang2000} using checkerboard. Besides the checkerboard pattern, there is also common a grid of circles \cite{heikkila1997four} to calibrate camera intrinsic. Some LiDAR intrinsic calibration methods are performed through a box \cite{atanacio2011lidar} or a planar wall \cite{muhammad2010calibration}. Extrinsic calibration between multiple sensors is commonly done by target-based methods, such as factory calibration, calibration room. Zhang et al. \cite{zhang2004} solve the extrinsic parameters based on a checkerboard, and refine them by minimizing the re-projection error of laser points to the checkerboard plane. Yan et al. \cite{2202.13708} propose a target-based joint calibration method by checkerboard with four round holes. Wang et al. \cite{wang2011integrating} perform radar-camera calibration through the metal panel. Per{\v{s}}i{\'c} et al. \cite{pervsic2017extrinsic} propose a complementary calibration target design suitable for both the LiDAR and the radar.

\subsection{Target-less Method}
In some cases, target-based methods are impractical, leading to the development of targetless methods. The targetless methods are more convenient than the target because no particular target is required. These methods use environmental features to match correspondences in sensor data. According to different information extraction methods, the research directions of multi-sensor online calibration can be divided into three methods: edge registration \cite{levinson2013}, mutual information \cite{pandey2012}, and segmentation \cite{zhu2020online, ma2021crlf}.
Barazzetti et al. \cite{barazzetti2011targetless} perform camera intrinsic calibration using only natural scenes.
Levinson et al. \cite{levinson2013} propose an online extrinsic calibration method by aligning edge features of objects in the environment. Some methods utilize the road features to perform automatic calibration of multi-sensors.
Ma et al. \cite{ma2021crlf} calibrate the extrinsic of the LiDAR and camera with line features in the road scene. Wei et al. \cite{2203.03182} calibrate the extrinsic of multiple LiDARs using road features in the road scene. Gong et al. \cite{gong20133d} propose a method in which they use trihedral target commonly found in urban and indoor environments.  
Pandey et al. \cite{pandey2015automatic} use the reflection intensity measured by LIDAR and the intensity value of the camera for the extrinsic calibration.
Similarly, Taylor et al. \cite{Taylor2012AutomaticCO} perform LIDAR-camera calibration by normalizing the mutual information, and maximizing the gradient correlation of image and LIDAR points.

\subsection{Motion-based Method}
Motion-based method treats  sensors extrinsic calibration as a  hand-eye calibration problem \cite{strobl2006optimal}. This method does not require field-of-view overlap between sensors, and  usually has better robustness to the initial value, but the accuracy of this method is low. As long as the sensor has odometer information, the problem can be transformed into a problem of solving a homogeneous system of linear equations \cite{shiu1987calibration}. There are different forms of solutions to this problem, such as quaternion form \cite{park1994robot}, dual quaternion form\cite{daniilidis1999hand}, and helical motion and helical axis forms\cite{fassi2005hand}. Different forms of solutions correspond to the same method, but compared with other forms of solutions that can only obtain rotation calibration, the dual quaternion form can additionally obtain translation calibration\cite{horn2021online}. However, these methods do not fully consider the uncertainty of measurement, resulting in the calibration accuracy being easily affected by sensor noise. Dornaika et al. \cite{dornaika1998simultaneous} unifies rotation and translation into a nonlinear optimization framework. Strobl et al. \cite{strobl2006optimal} introduces the Euclidean group, which takes into account the relative errors of the rotation and translation. Recent work such as Huang et al. \cite{huang2017extrinsic} apply the Gauss-Helmert model to the solution of motion constraints, and simultaneously optimize the relative motion estimation and external parameters of the sensor, making it highly robust to sensor noise and initial values of external parameters.

Temporal calibration is also an important component of the motion-based method. \cite{rehder2016general} Generally, there are three methods of temporal calibration, including hardware synchronization, software synchronization with bidirectiona  communications and software synchronization with unidirectional communications. Our temporal calibration belongs to the third method, which is finished in two ways, using the odometry data of each sensor, obtaining temporal calibration through an optimized solution, or obtaining calibration results through data correlation. Paul et al. \cite{furgale2013unified} propose a unified method of determining fixed time
offsets between sensors using batch, continuous-time, maximum-likelihood estimation. Qin et al. \cite{qin2018online} propose a new IMU and camera temporal calibration algorithm. Different from other algorithms that compensate for the camera pose or IMU pose by multiplying the speed and angular velocity by the time difference. This algorithm assumes that the feature points move on the image plane at a constant speed in a short time, delaying the time between the IMU and the camera, converting to delays of feature locations on the image plane, simplifying the residual function. Qiu et al. \cite{qiu2020real} proposed a new temporal calibration method through motion correlation analysis.

\subsection{Learning-based Method}
In recent years, deep learning networks have shown effectiveness in dealing with both 2D and 3D computer vision tasks such as localization, object detection, and segmentation. There are also a few works that apply deep learning to sensor calibration tasks, especially adapting to camera and LiDAR calibration problems. 
Schneider et al. \cite{DBLP:journals/corr/SchneiderPSF17} propose the first deep convolution neural network RegNet for LiDAR-camera calibration that infers extrinsic parameters directly, by regressing to calibration parameters based on a supervised network. To further involve spatial information and eliminate the influence of intrinsic, Iyer et al. \cite{2018calibnet} propose a geometrically supervised network CalibNet for LiDAR-camera calibration by reducing the dense photometric error and dense point cloud distance error instead. RGGNet \cite{RGGnet2020} considers Riemannian geometry and utilizes a deep generative model to learn an implicit tolerance model for LiDAR-camera calibration. In comparison, there are much fewer learning-based approaches for intrinsic calibration problems. DeepCalib \cite{bogdan2018deepcalib} proposes a CNN-based approach for intrinsic calibration of wide FOV cameras with an automatically generated large-scale intrinsic dataset. Because of less intuitive spatial relations in camera intrinsic calibration problems and high-precision requirements, learning-based methods have shown fewer advantage. 
Aside from end-to-end deep learning network for sensor calibration, some work on relevant vision tasks can be adapted as key partitions of the calibration process, such as vanishing point detection \cite{deepvp2018}, car heading prediction \cite{MonoEF2021} and camera pose estimation \cite{2017demon, kendall2016posenet,teed2020deepv2d, GeoNet2018, zhou2017unsupervised}.
Although learning-based methods leverage the strength of deep learning for spatial feature extraction and matching more effectively without human interaction after model training, these approaches are limited to calibration for certain sensors, particularly cameras and LiDAR. And compared to non-learning methods, most learning-based approaches are less stable and less precise, while precision is crucial in calibration tasks. So, in general, deep network is not commonly devoted for multi-sensor calibration problems.

\section{METHODOLOGY}
In this section, we introduce the details of our calibration toolbox, including manual calibration tools, automatic calibration tools, factory calibration tools, and online calibration tools.

\subsection{Manual Target-less Calibration Tools}
Manual calibration known as hand calibration is the simplest type of calibration methods for autonomous vehicles. Though the principles and the accomplishment process is fairly simple compared to other methods, the user can get a calibration result of high accuracy and reliability if they spend enough time on those methods. Early manual calibration methods rely on target-like checkerboards or ordinary boxes, the calibration procedure is often carried out in customized calibration room. However, this complicated setup is costly for small autonomous driving companies and individual users. 

\begin{figure}[h]
\centering 
\includegraphics[width=0.45\textwidth]{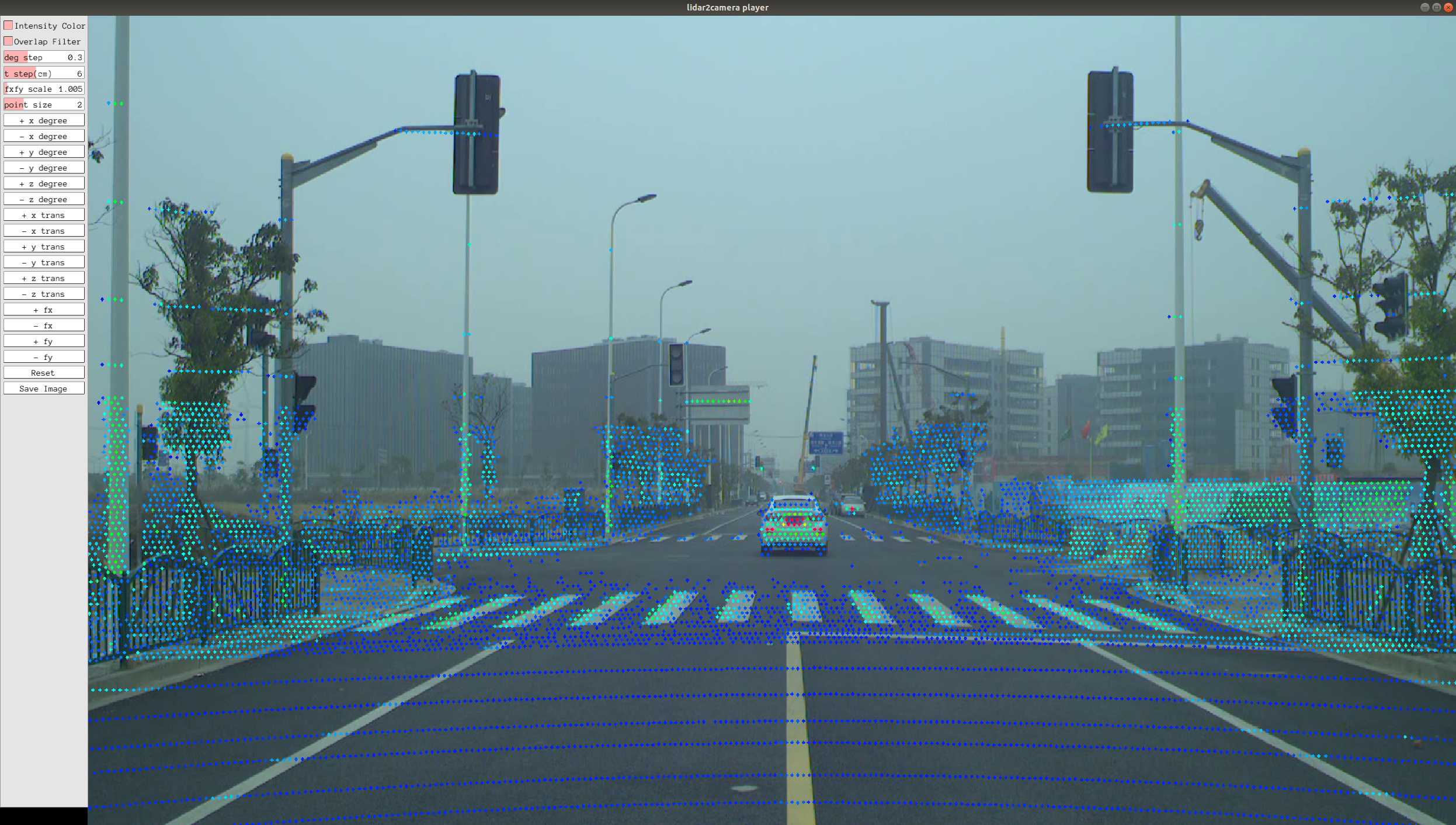} 
\caption{LiDAR2Camera calibration tool interface} 
\label{Fig.lidar2camera} 
\end{figure}

Therefore, our calibration toolbox provides users with four manual calibration tools for autonomous vehicles in target-less and arbitrary road scenes. Obviously, a road scenes with strong and well-recognized features like trees and traffic signs will lead to a better calibration result. The extrinsic parameters can be adjusted in the user-friendly control panel on the left shown in Fig. \ref{Fig.lidar2camera}, or keyboards shown in Table \ref{tab:lidar2camera} and Table \ref{tab:lidar2camera1} can serve as the control input to achieve the calibration, and examples of well-calibrated scenes can be seen in Fig. \ref{Fig.lidar2camera}. The four calibration tools will be introduced in detail in the following part.

\subsubsection{LiDAR to camera calibration}
The whole calibration process is to align a 3D LiDAR point cloud and an image in road scenes by tuning the intrinsic and extrinsic parameters. The 3D point cloud is denoted as $\mathbf{p}_i^\mathrm{L}$$=$$(Xi,Yi,Zi)^T$$\in$$\mathbb{R}^3$ in LiDAR coordinate frame, then the point can be transformed to camera frame $\mathbf{p}_i^\mathrm{C}$$=$$(Xc,Yc,Zc)^T$$\in$$\mathbb{R}^3$ according to
\vspace{-2mm}
\begin{equation}
\label{equ:extrinsic}
    \mathbf{p}_i^\mathrm{C} = \boldsymbol{\mathrm{R}} \cdot \mathbf{p}_i^\mathrm{L} + \boldsymbol{\mathrm{t}},
\vspace{-2mm}
\end{equation}
where R represents the rotation parameters and t represents the translation parameters, together form the extrinsic parameter. Next, $\mathbf{p}_i^\mathrm{C}$ is projected onto the image plane through a projection function:
\vspace{-2mm}
\begin{equation}
\label{equ:project}
    \boldsymbol{\mathcal{K}}: \mathbb{R}^3 \rightarrow \mathbb{R}^2, \mathbf{q}_i = \boldsymbol{\mathcal{K}} (\mathbf{p}_i^\mathrm{C}) \ .
\vspace{-2mm}
\end{equation}
$\boldsymbol{\mathcal{K}}$ is the intrinsic parameter of the camera, which is defined by the camera characteristics such as focal length and lens distortion, and is calibrated by Zhang's method \cite{zhang2000}. Beside these parameters, the update step of adjustment and point size can also be tuning in the panel. It is worth mentioning that the $ Intensity Color$ button can change the display mode to intensity map display mode, and the $Overlap Filter$ button is to eliminate overlap LiDAR points within a depth of 0.4m.

\begin{table}[]
\centering
\caption{Keyboard commands for extrinsic parameter adjustment }
\vspace{-3mm}
\label{tab:lidar2camera}
\begin{tabular}{|c|c|c|c|}
\hline
Extrinsic Params & Keyboard Input & Extrinsic Params & Keyboard Input \\ \hline
$+x (degree)$ & $q$ & $-x (degree)$ & $a$ \\ \hline
$+y (degree)$ & $w$ & $-y (degree)$ & $s$ \\ \hline
$+z (degree)$ & $e$ & $-z (degree)$ & $d$ \\ \hline
$+x (trans)$ & $r$ & $-x (trans)$ & $f$ \\ \hline
$+y (trans)$ & $t$ & $-y (trans)$ & $g$ \\ \hline
$+z (trans)$ & $y$ & $-z (trans)$ & $h$ \\ \hline
\end{tabular}
\end{table}

\begin{table}[]
\centering
\caption{Keyboard commands for intrinsic parameter adjustment }
\vspace{-3mm}
\label{tab:lidar2camera1}
\begin{tabular}{|c|c|c|c|}
\hline
Intrinsic Params & Keyboard Input & Intrinsic Params & Keyboard Input \\ \hline
$+fy (degree)$ & $i$ & $-x (degree)$ & $k$ \\ \hline
$+fx (degree)$ & $u$ & $-y (degree)$ & $j$ \\ \hline
\end{tabular}
\end{table}

\subsubsection{LiDAR to LiDAR calibration}
This calibration process is to achieve 3-D point cloud registration between source and target point cloud by  adjusting only the extrinsic parameters. The source point cloud is denoted as $\mathbf{p}_i^\mathrm{S}$$=$$(Xi,Yi,Zi)^T$$\in$$\mathbb{R}^3$ while the target point cloud is denoted as  $\mathbf{p}_i^\mathrm{T}$$=$$(Xi,Yi,Zi)^T$$\in$$\mathbb{R}^3$, then the rigid transformation between two clouds can be represented as:
\vspace{-2mm}
\begin{equation}
\label{equ:extrinsic}
    \mathbf{p}_i^\mathrm{S} = \boldsymbol{\mathrm{R}} \cdot \mathbf{p}_i^\mathrm{T} + \boldsymbol{\mathrm{t}}
\vspace{-2mm}
\end{equation}

\begin{figure}[h]
\centering 
\includegraphics[width=0.45\textwidth]{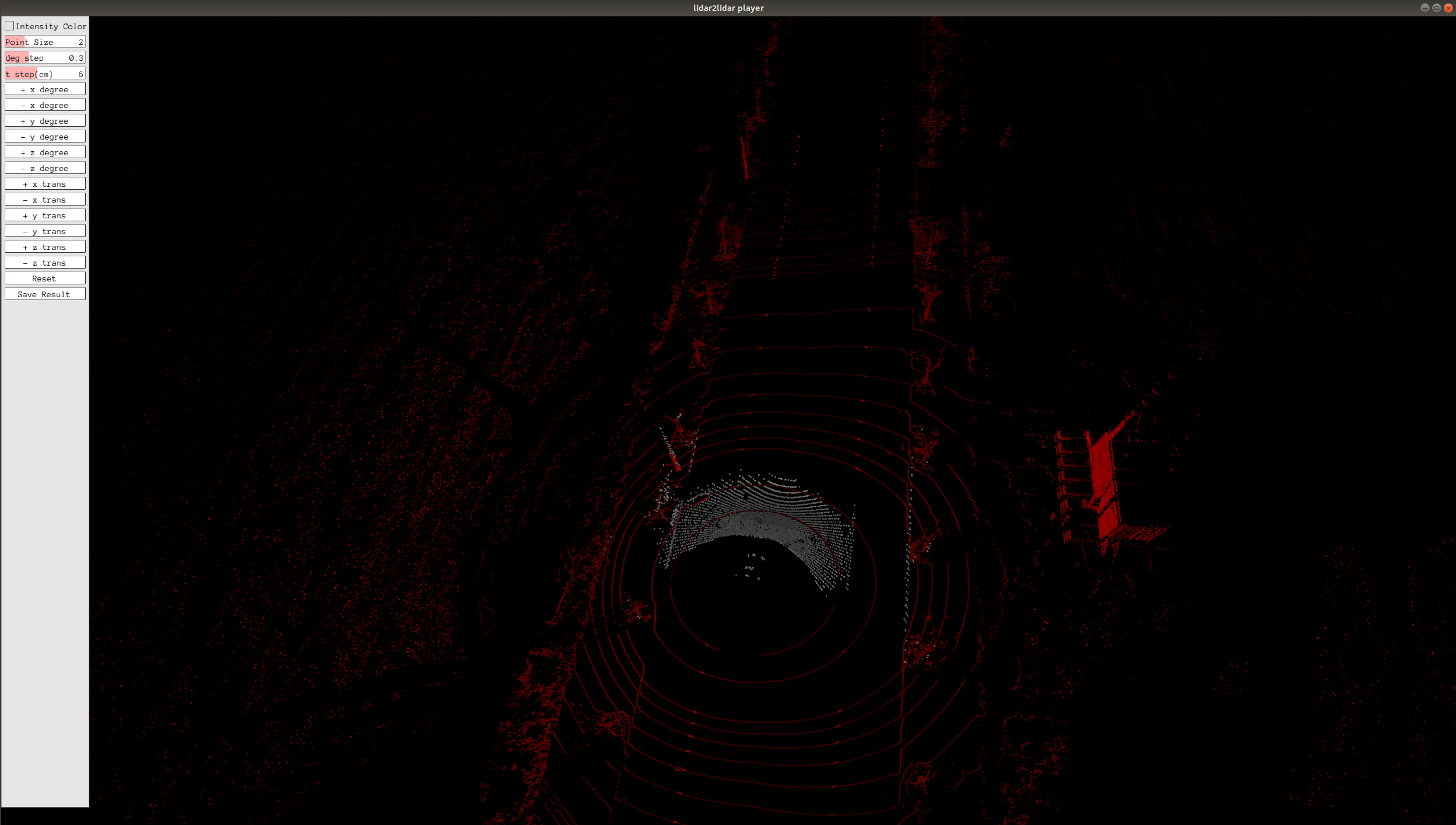} 
\caption{LiDAR2LiDAR calibration tool interface} 
\label{Fig.lidar2lidar} 
\end{figure}

\subsubsection{Radar to LiDAR calibration}
The calibration process is almost the same to LiDAR-to-LiDAR Calibration, but does the registration between 2-D radar point cloud and 3-D LiDAR point cloud. Therefore, there will be less parameters to be tuned, one degree lost for R and t.

\begin{figure}[h]
\centering 
\includegraphics[width=0.45\textwidth]{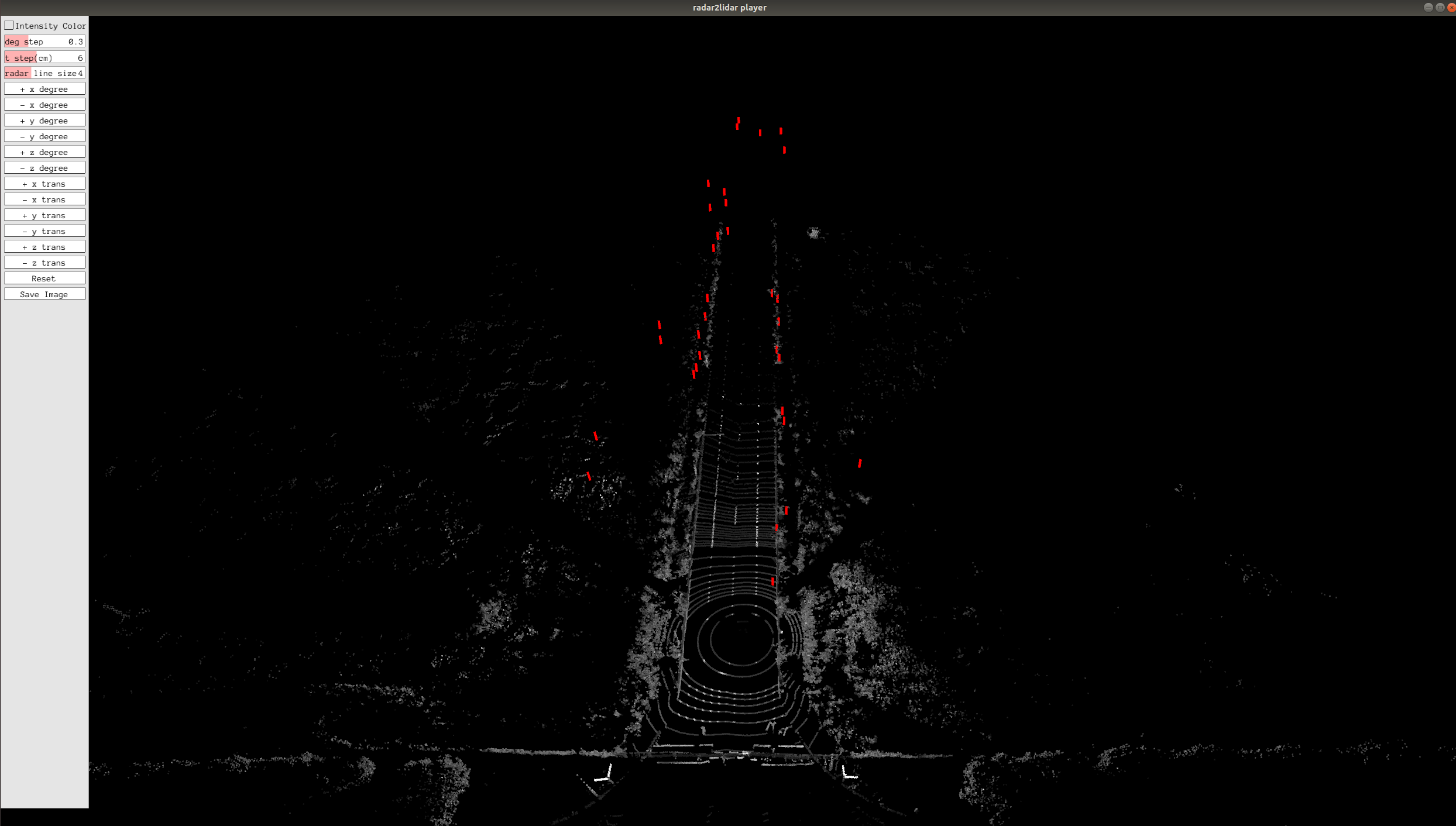} 
\caption{Radar2LiDAR calibration tool interface} 
\label{Fig.radar2lidar} 
\end{figure}

\subsubsection{Radar to camera calibration}

\begin{figure}[h]
\centering 
\includegraphics[width=0.45\textwidth]{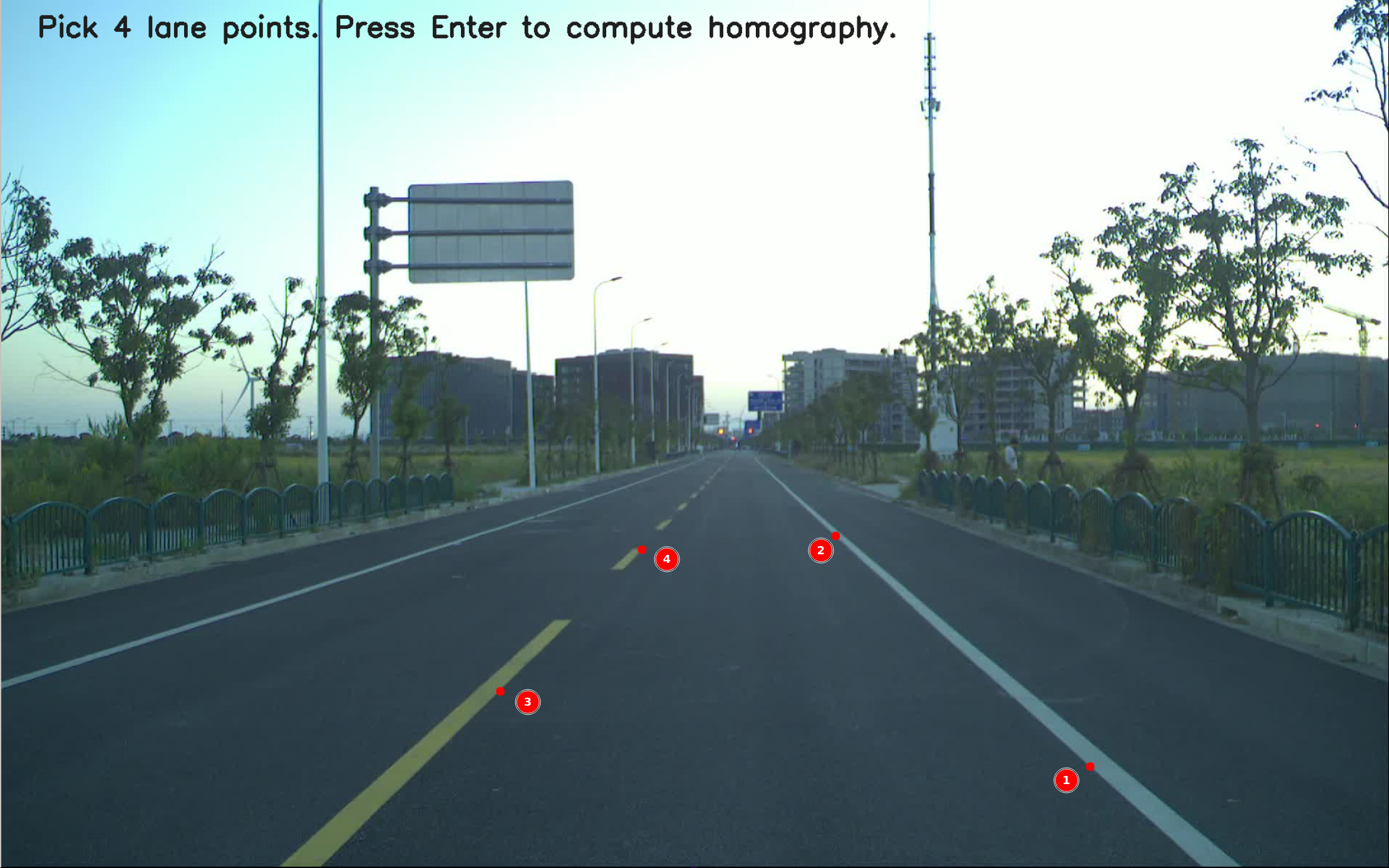} 
\caption{Radar2Camera calibration tool interface} 
\label{Fig.radar2camera_pick} 
\end{figure}
\begin{figure}[h]
\centering 
\includegraphics[width=0.45\textwidth]{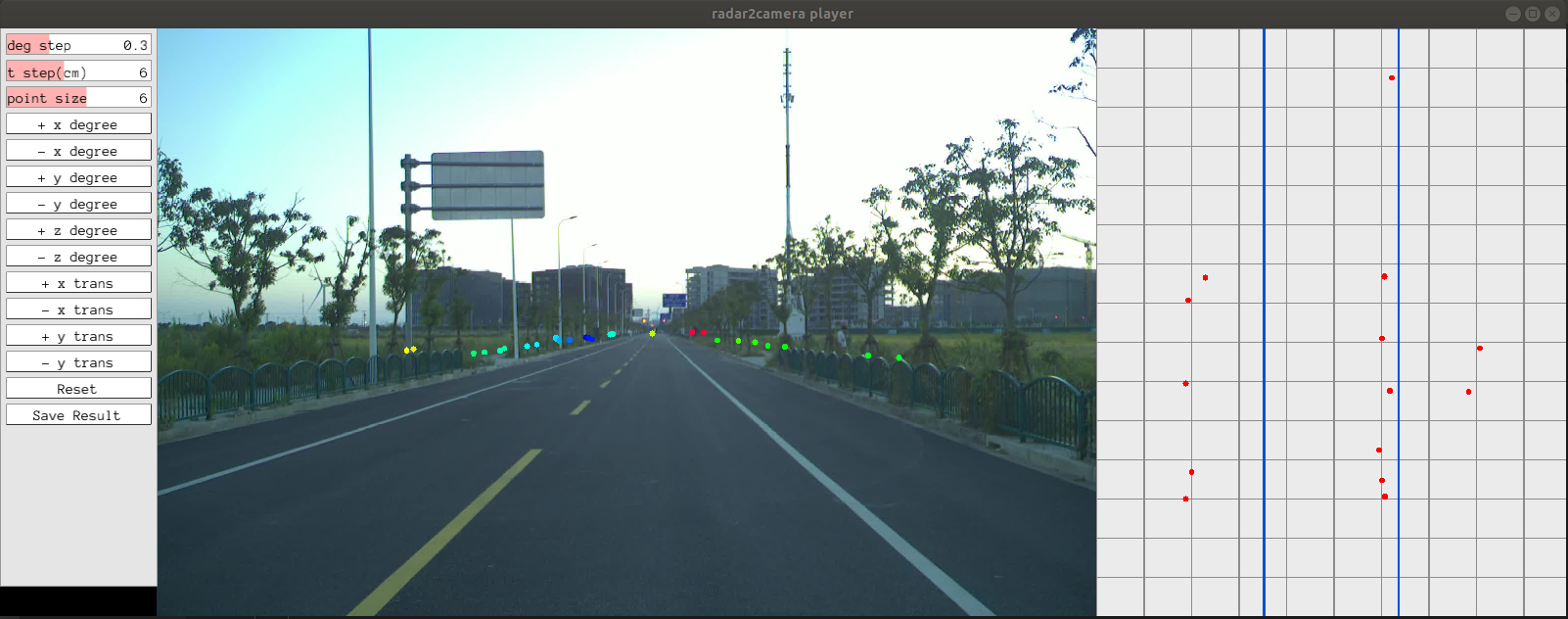} 
\caption{Radar2Camera calibration tool interface} 
\label{Fig.radar2camera} 
\end{figure}

For the calibration of Radar and camera, because Radar is a two-dimensional device, we try to keep it parallel to the ground during installation. For the camera, we need to find the homography matrix of its correspondence with the ground. Then we visualize it on the image and on the bird's-eye view, and when the alignment is displayed on the two images, we can consider the calibration complete. Fig.\ref{Fig.radar2camera_pick} shows how the camera-to-ground homography matrix is calculated, selecting two points each in the left and right lanes. Fig.\ref{Fig.radar2camera} shows the projection results of Radar in the image and the bird's-eye view. The parallel lines in the bird's eye view indicate the direction of the lane line.

\subsection{Automatic Target-based Calibration Tools}
\subsubsection{Camera calibration}
The camera calibration method commonly used in the industry is Zhang's method \cite{zhang2000}. Zhang's method extracts the corner points of the checkerboard through the calibration board placed in various poses and calculates the intrinsic and distortion parameters of the camera. The pinhole model is usually
used when calibrating the camera’s intrinsic, 
but the pinhole model is just a simplified model of the camera projection process \cite{stu2006img}. The actual camera lens group is more complex and does not have an absolute optical center point \cite{juarez2020distorted}. Due to the complex internal structure of the camera, lacking an effective quantitative evaluation method for the camera’s intrinsic calibration.

Since the degree of distortion of each camera lens is different, this lens distortion can be corrected by camera calibration to generate a rectified image. The rectified image is crucial for subsequent perception or joint calibration with other sensors. Therefore, it is necessary to develop a quantitative evaluation method of distortion to ensure the accuracy of distortion calibration.
We developed a program for the quantitative evaluation of camera distortion based on Tang's method \cite{tang2012high}.
The method is an easy-to-implement and automated procedure to perform camera distortion evaluation.

\begin{figure}[h]
\centering 
\includegraphics[width=0.5\textwidth]{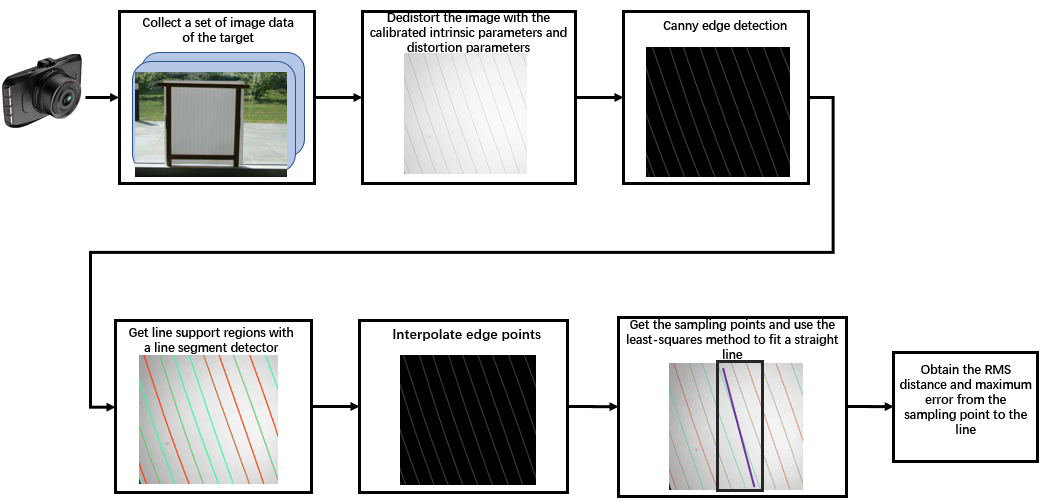} 
\caption{ evaluation of camera distortion process.} 
\label{Fig.evalu} 
\end{figure}

The target for distortion evaluation is composed of fishing lines and translucent paper. Distortion parameter evaluation flow charts as shown in Fig.\ref{Fig.evalu}. 
First, the camera is used to capture an image of the target, and the target fills the entire image. This image is then de-distorted by the camera calibration parameters to obtain the rectified image. Then, the linear segment detection algorithm uses the Canny descriptor \cite{canny1986computational} to extract straight line segments in the rectified image. Because of the use of NMS (Non-Maximum suppression), the linear segments are discontinuous, and linear interpolation is used to obtain a continuous straight line. After Gaussian sampling, many linear sampling points are obtained. Finally, the fitted straight line is obtained by the least squares method. According to formula \eqref{equ:RMS} and formula \eqref{equ:d_max}, the root mean square distance and the maximum error distance from the sampling point to the fitted straight line are obtained. The regression line can be expressed as ${\alpha}*x + {\beta}*y - {\gamma} = 0$, Suppose given $L$ straight lines.

\vspace{-2mm}
\begin{equation}
\label{equ:RMS}
    S = \sum_{i=1}^L \sum_{i=1}^{N_l} |S_{li}|^2 =\sum_{i=1}^L \sum_{i=1}^{N_l} (\mathbf{\alpha}_l*x_{li} + \mathbf{\beta}_l*y_{li} - \mathbf{\gamma}_l)^2
\vspace{2mm}
\end{equation}

\vspace{-2mm}
\begin{equation}
\label{equ:d_max}
    d_{max} = \sqrt{((\sum_{l=1}^L |max_i S_{li} - min_i S_{li}|^2)/L)}
\vspace{-2mm}
\end{equation}

The root mean square distance and maximum distance are measurement indicators for evaluating the quality of distortion parameters.

\subsubsection{LiDAR to camera calibration}
Target-based calibration methods for LiDAR and camera rely on observing artificial calibration targets placed in front of the sensor system, simultaneously by both the sensors to obtain the position of the feature point. For the calibration of LiDAR and camera, the existing method is generally to calibrate the intrinsic of the camera first and then calibrate the extrinsic of the LiDAR and camera. If the camera’s intrinsic is not calibrated correctly in the first stage, it isn’t easy to calibrate the LiDAR-camera extrinsic accurately. In order to solve the influence of the inaccurate calibration of the camera’s intrinsic in the existing method on the extrinsic parameters from LiDAR to camera. We proposed a joint calibration method. If you want to know the specific details, please refer to our previous work \cite{2202.13708}.

\begin{figure}[ht]
\centering
\includegraphics[scale=0.24]{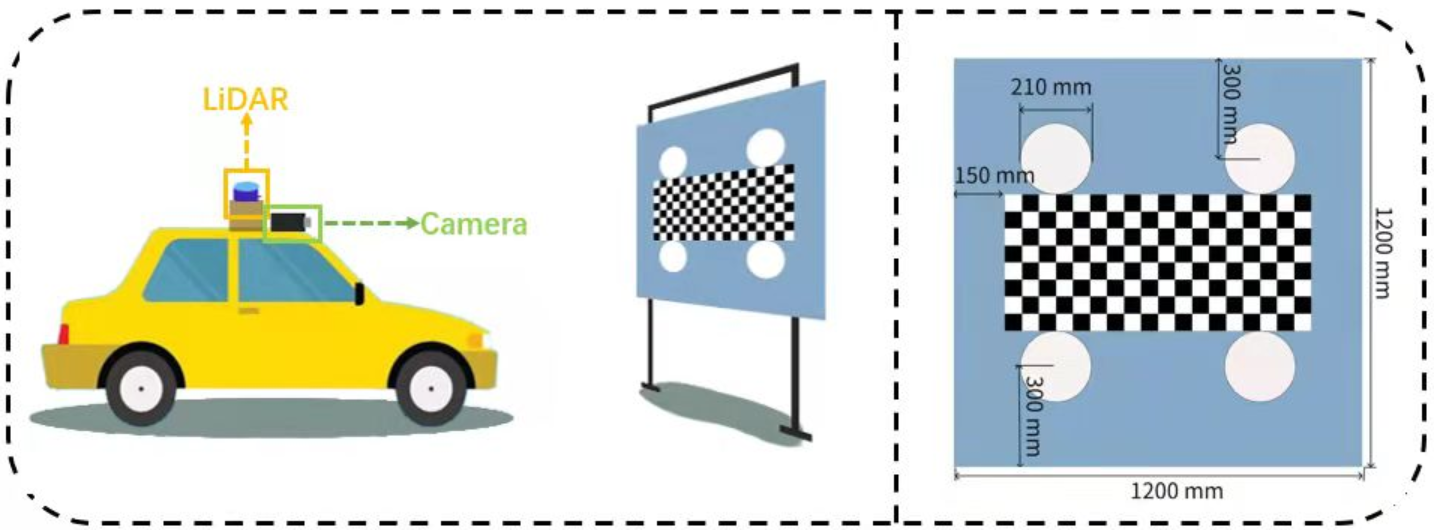}
\caption{A novel calibration board pattern.} \label{fig:carboard}
\end{figure}

We designed a novel calibration board pattern, as shown in Fig.\ref{fig:carboard}, which contains a checkerboard used for the calibration of the camera’s intrinsic parameters and several circular holes for locating the LiDAR point cloud. We first calibrate the camera initial intrinsic and board-camera initial extrinsic parameter by Zhang’s method\cite{zhang2000}. Then, 2D circles center points on the image are calculated from these parameters and the calibration board size. By extracting the position of the circles center in LiDAR, we can project the circles center 3D points to the image plane by the LiDAR-camera calibration parameters. The calculated 2D points and the projected 2D points form multiple 2D point pairs. We use the Euclidean distance between these point pairs to refine the calibration parameters. At the same time, the constraints on reprojection of 3D-2D points of checkerboard corners are added to the optimization process.

The whole calibration process aims to achieve better alignment between LiDAR point cloud and image with satisfying the checkerboard corner constraints, which is also considered one of the critical evaluation criteria. The number of checkerboard corners on our calibration board is much more than the number of circular holes. Therefore, the weight of LiDAR-camera alignment error is greater than that of board-camera alignment error to balance the effects of two-loss functions. In sum, we minimize the following objective function:

\vspace{-2mm}
\begin{equation}
\label{equ:extrinsic}
    \boldsymbol{\mathcal{J}_{sum}} = \alpha {\boldsymbol{\mathcal{J}_{board}}} + \beta\boldsymbol{\mathcal{J}_{lidar}} 
\vspace{-2mm}
\end{equation}
where $\mathcal{J}_{board}$ represents the checkerboard corners reprojection
error of the calibration board. $\mathcal{J}_{lidar}$ represents The circles center reprojection error of the calibration board. In  our  experiment, $\alpha$ is set to 1, and $\beta$ is set to 60. $\mathcal{J}_{board}$ and $\mathcal{J}_{lidar}$ can be calculated by the following formulas.

\begin{equation}
\begin{aligned}
    \boldsymbol{\mathcal{J}_{board}}&= \sum\limits_{(u,v) \in P_{B}}(||u - u_{det}||_2 + ||v - v_{det}||_2)
\end{aligned}
\label{}
\end{equation}
where $(u,v)$ represents the pixel coordinate point of the $P_B$ projection, $(u_{det}, v_{det})$ is the actual detected pixel point.

\begin{footnotesize} 
\begin{equation}
\begin{aligned}
    \boldsymbol{\mathcal{J}_{lidar}}&= \sum\limits_{(u,v) \in P_{L}}(||u - u_{det}||_2 + ||v - v_{det}||_2)
\end{aligned}
\end{equation}
\end{footnotesize}
where $(u,v)$ represents the pixel coordinate point of the $P_L$ projection,  $(u_{det}, v_{det})$ is the actual calculated pixel point.

\subsection{Automatic Target-less Calibration Tools}

\subsubsection{IMU heading calibration}
IMU heading calibration aims to correct the installation error of forwarding orientation between IMU and vehicle. So, we only calibrate the yaw angle offset of IMU to align the orientation, denoted as $\mathbf{\gamma}_\mathrm{offset}$. We derive the direction of the vehicle from the driving route for each timestamp as $\mathbf{\gamma}_\mathrm{gd}$. And the offset between estimated driving orientation and measured IMU yaw angle $\mathbf{\gamma}_\mathrm{IMU}$ is the calibration result.

\begin{figure}[h]
\centering 
\includegraphics[width=0.5\textwidth]{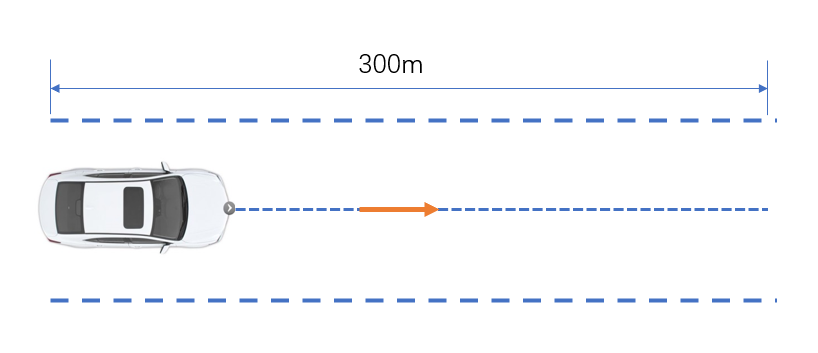} 
\caption{A straight line data for IMU heading calibration.} 
\label{Fig.imu_heading} 
\end{figure}

We apply the b-spline method to smooth the driving route based on sensor localization data. Besides, not all data in one driving trip is used, only the straight driving route is picked out for the following calibration. By removing data with a rapid change of driving orientation, such as a U-turn, we can get a precise approximation of real-time true yaw angle. The calibration formula can be  described as
\vspace{-2mm}
\begin{equation}
\label{equ:extrinsic}
    \mathbf{\gamma}_\mathrm{offset} = \frac{1}{|S_l|}\sum_{l\in S_l}(\mathbf{\gamma}_\mathrm{gd}^{l} - \mathbf{\gamma}_\mathrm{IMU}^{l})
\vspace{-2mm}
\end{equation}
where $S_l$ is the set of straight paths. Of course, in order to improve the accuracy of calibration, we suggest that you can directly record a straight line data for calibration, as shown in Fig.~\ref{Fig.imu_heading}.

\subsubsection{LiDAR to camera calibration}
Accurate calibration of LiDAR and camera is one of the most common calibrations in autonomous driving. Monocular vision-based perception systems achieve satisfactory performance at a low cost but do not provide reliable 3D geometric information. Camera-LiDAR fusion perception is to improve performance and reliability. The premise and assumption of camera and LiDAR fusion is accurate calibration between camera and LiDAR, including camera intrinsic and camera and LiDAR extrinsic parameters.
After the LiDAR and camera are accurately calibrated, the calibrated parameters will gradually become inaccurate due to the long-term movement of the vehicle and the influence of other factors such as temperature. Since the perceptual fusion algorithm is very sensitive to the accuracy of the calibration parameters, this can seriously degrade the performance and reliability of the perceptual fusion algorithm. At this point, re-calibration through a calibration room or manually is cumbersome and impractical, so we developed a tool to automatically calibrate the LiDAR and camera in road scenes. The specific details of this method can be found in our previous work \cite{ma2021crlf}. This work proposes a method for the extrinsic parameter calibration of LiDAR and camera in common road scenes. The method is completely automated, and its efficiency and accuracy are also relatively high.

\begin{figure}[h]
\centering 
\includegraphics[width=0.5\textwidth]{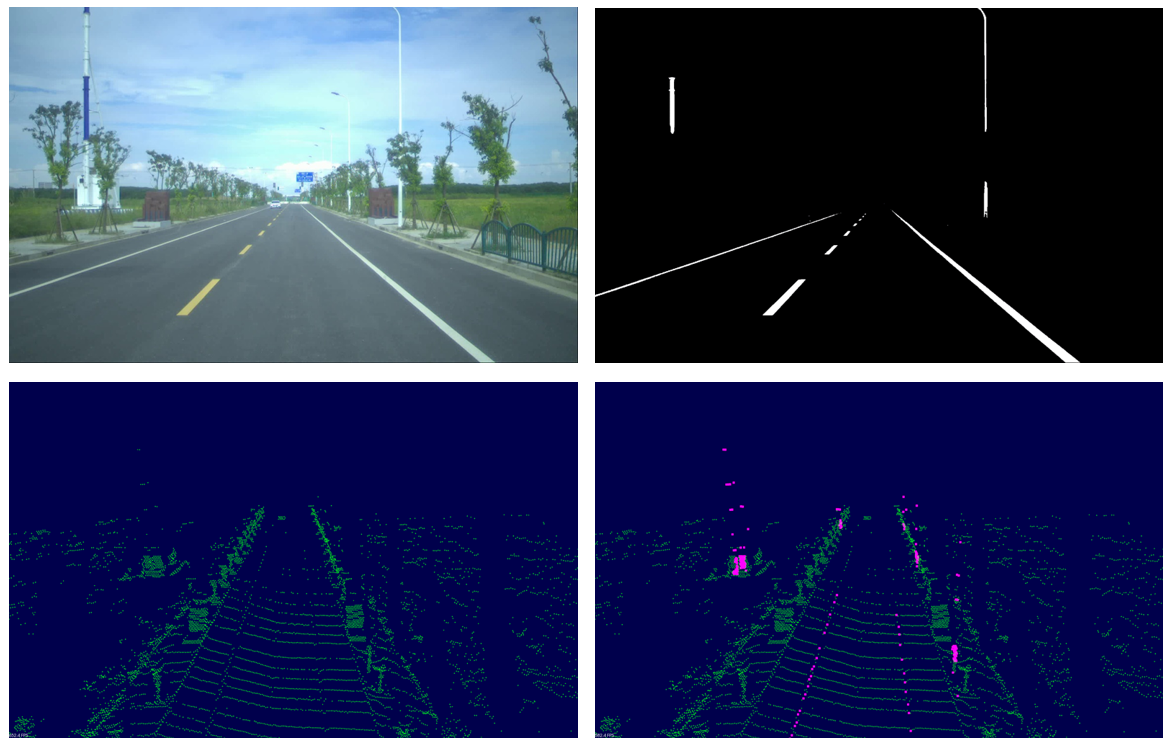} 
\caption{Image and point cloud segmentation process.} 
\label{Fig.lidar2camera_segmentation} 
\end{figure}

\begin{figure}[h]
\centering 
\includegraphics[width=0.5\textwidth]{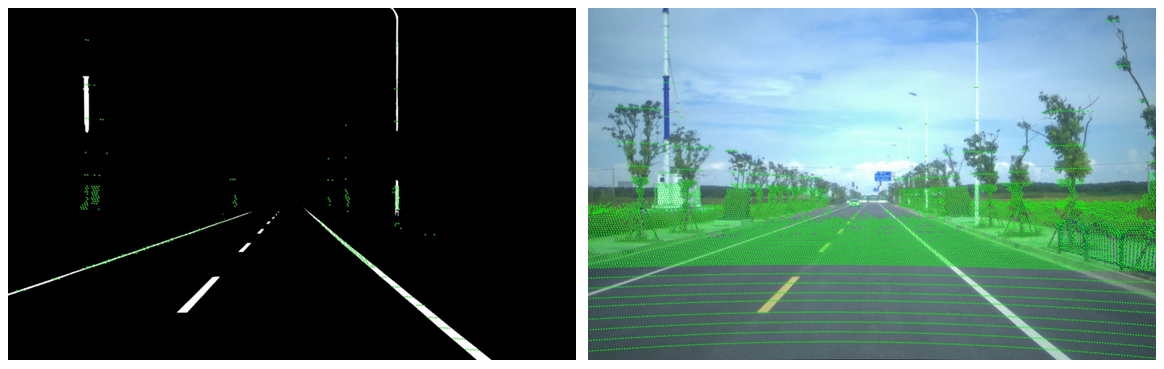} 
\caption{Mask and image calibration projection results.} 
\label{Fig.lidar2camera_calib} 
\end{figure}

First, a set of road scene data is collected by LiDAR and camera, and then linear features such as lane lines and road signs are extracted from the image and point cloud, respectively. As shown in Fig.~\ref{Fig.lidar2camera_segmentation}, we extract lane lines and road poles from images by BiSeNet-V2\cite{yu2018bisenet} and from point clouds by intensity and geometry methods. Subsequently, a cost function is designed to refine the initial extrinsic calibration parameters and ensure the error lies within an acceptable range. The pixels from poles $Q_{pole}$ and from road lanes $Q_{lane}$ can be directly obtained from the class label ``pole" and ``road lane". Combining the segmentation results, we can obtain two binary masks $\boldsymbol{\mathcal{M}}_{line}: \mathbb{R}^2 \rightarrow \{0,1\}, line\in\{pole,lane\}$  on the pixel coordinate defined as follows: 
\vspace{-2mm}
\begin{equation} 
\begin{split}
    \boldsymbol{\mathcal{M}}_{line}(\mathbf{q}) :=
    \begin{cases}
        1& \mathbf{q}\in Q_{line}\\
        0& \mathrm{otherwise}
    \end{cases}.
\end{split}
\vspace{-2mm}
\end{equation} 
After line features are extracted from both the image and the point cloud, we propose several cost functions that measure how well the image and the point cloud is correlated given an extrinsic parameter $(\mathbf{r},\mathbf{t})$.  Similar to \cite{zhu2020online}, we apply an inverse distance transformation (IDT) to the mask $ \boldsymbol{\mathcal{M}}_{line}$ to avoid duplicate local maxima during later optimization. The resulting height map  $\boldsymbol{\mathcal{H}}_{line}, line\in\{pole,lane\}$ is defined as follows:
\vspace{-2mm}
\begin{equation}
      \boldsymbol{\mathcal{H}}_{line}(\mathbf{q}) :=
    \begin{cases}
\max\limits_{\mathbf{s} \in \mathbb{R}^2 \setminus Q_{line}} \gamma_0^{\lVert \mathbf{q}-\mathbf{s} \Vert_1}& \mathbf{q}\in Q_{line}\\
0 & \mathbf{q}\in \mathbb{R}^2 \setminus Q_{line}
\end{cases}  .
\vspace{-2mm}
\end{equation}
Next, we propose the projection cost function $ \boldsymbol{\mathcal{J}_{Proj}}: (\mathbf{r},\mathbf{t}) \rightarrow \mathbb{R}$, which represents the consistency between the projected pixels of $P_{lane}$ and $P_{pole}$ and their corresponding masks in the image. And $ \boldsymbol{\mathcal{J}_{Proj}}$ is defined as
\vspace{-2mm}
\begin{equation}
\footnotesize
\begin{split}
    \boldsymbol{\mathcal{J}_{proj}}   =  \tanh( \tau_1 \sum\limits_{line \in \{pole, \ lane \} }  & \displaystyle\frac{\sum\limits_{\mathbf{p} \in P_{line}^\mathrm{L} } \boldsymbol{\mathcal{H}}_{line}\circ \boldsymbol{\mathcal{K}}(\mathbf{R}(\mathbf{r}) \mathbf{p} +\mathbf{t})}{|P_{line}^\mathrm{L} |})  
\end{split},
\vspace{-3mm}
\end{equation}
The notation $\circ$ refers to obtain the height value with projected pixel location.
$|P_{line}^\mathrm{L} |$ is the number of points in $P_{line}^\mathrm{L}$  and is used to balance the cost between poles and lanes. The larger this cost function is, the better the semantic features from two data domains match. Fig.~\ref{Fig.lidar2camera_calib} shows the calibration projection results of the mask and the image. 

\subsubsection{LiDAR to IMU calibration}
The calibration of LiDAR and IMU is also one of the common calibrations for autonomous driving, and its calibration accuracy has a great impact on LiDAR mapping and positioning modules. Generally, when calibrating the extrinsic parameters of LiDAR to IMU, the calibration accuracy is judged by judging whether the local mapping of the LiDAR is good. The calibration process is to solve the extrinsic parameters from the LiDAR to the IMU through the local map constructed by the sliding window. This calibration tool was developed based on \cite{BALM2021}. 
The feature points are distributed on the same edge line or plane on the local map by minimizing the eigenvalues of the covariance matrix. The method minimizes the sum of distances from feature points to feature planes or edge lines by minimizing the eigenvalues of the covariance matrix and optimizes to achieve the purpose of extrinsic parameter calibration from LiDAR to IMU. The BA algorithm \cite{BALM2021} minimizes the distance from each plane feature point to the plane. The formula is as follows. 
\vspace{-1mm}
\begin{equation}
\begin{aligned}
\label{equ:extrinsic}
    \mathbf{(T^*, n^*, q^*)} &= 
    \mathop{argmin}\limits_{\mathbf{T,n,q}}\frac{1}{N}\sum_{i=1}^{N}\mathbf{(n^T(p_i - q))^2}\\
    &=\mathop{argmin}\limits_{\mathbf{T}}(\mathop{min}\limits_{\mathbf{n,q}}\sum_{i=1}^{N}\mathbf{(n^T(p_i - q))^2}
\vspace{-1mm}
\end{aligned}
\end{equation}
$P_i$ is the point at which the window frame to be optimized is projected onto the local map. $q$ is a point on some feature (edge line, in-plane). $n$ is the plane normal vector. 
The optimized feature point position and the feature normal vector (direction vector) can be written as a function of the pose $T$, so only the pose $T$ needs to be optimized. Adaptive voxels can speed up search feature extraction. That is, iteratively divide the grid from $1m$, knowing that the points in the grid belong to the same edge line or the plane is unknown. In this method, the pose of the LiDAR in the world coordinate system at any time $t$ can be obtained, and the required rough initial extrinsic parameters can be obtained by the following formula.
\vspace{-1mm}
\begin{equation}
\label{equ:extrinsic}
    \mathbf{R}_I^\mathrm{L}, \mathbf{P}_I^\mathrm{L} = 
    \mathop{argmin}\limits_{\mathbf{R}_I^\mathrm{L},\mathbf{P}_I^\mathrm{L}}(||\mathbf{R}_I^\mathrm{L} \mathbf{T(t)}_I^\mathrm{M} - \mathbf{P}_I^\mathrm{L} - \mathbf{T(t)}_L^\mathrm{M}||^2)
\vspace{-1mm}
\end{equation}
Since the influence of the offset on the system will gradually increase with the increase of time, we use local mapping optimization to further correct it. The first frame in the sliding window is denoted as the $P$ frame, and the current frame is denoted as the $O$ frame. It is known that the pose transformation relationship from the IMU coordinate system to the world coordinate system at time $P$ and time $O$ is ${T}_{I_p}^W$ and ${T}_{I_o}^W$, and the transformation matrix from LiDAR to IMU denoted as ${T}_{L}^I$, according to the chain rule of coordinates, the transformation relationship between the LiDAR coordinate system at time $P$ and the LiDAR coordinate system at time $O$ can be obtained as shown in the following formula.
\vspace{-1mm}
\begin{equation}
\label{equ:extrinsic}
    \mathbf{T}_P^\mathrm{O} = \mathbf{T}_L^\mathrm{I} \mathbf{T}_{I_o}^W \mathbf{T}_{I_P}^{W-1} \mathbf{T}_{L}^{I-1}
\vspace{-1mm}
\end{equation}
Build a local map from this formula. After the local map is established, the corresponding relationship between each laser point cloud frame and the local map from frame $O$ to subsequent frames is further searched. 
\begin{figure}[h]
\centering 
\includegraphics[width=0.5\textwidth]{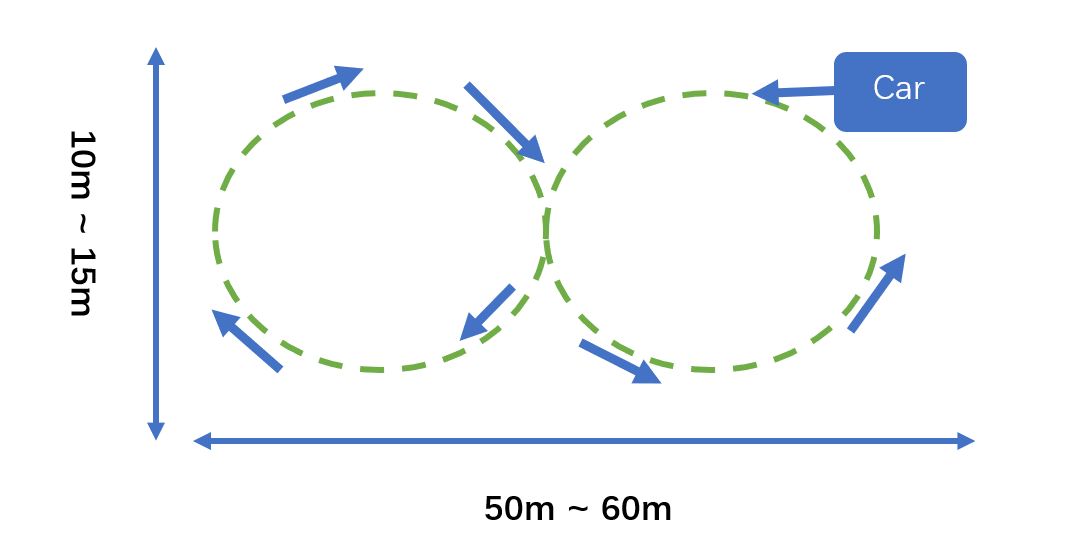} 
\caption{Driving along the 8-character trajectory for LiDAR to IMU calibration.} 
\label{Fig.lidar2imu} 
\end{figure}

\begin{figure}[h]
\centering 
\includegraphics[width=0.5\textwidth]{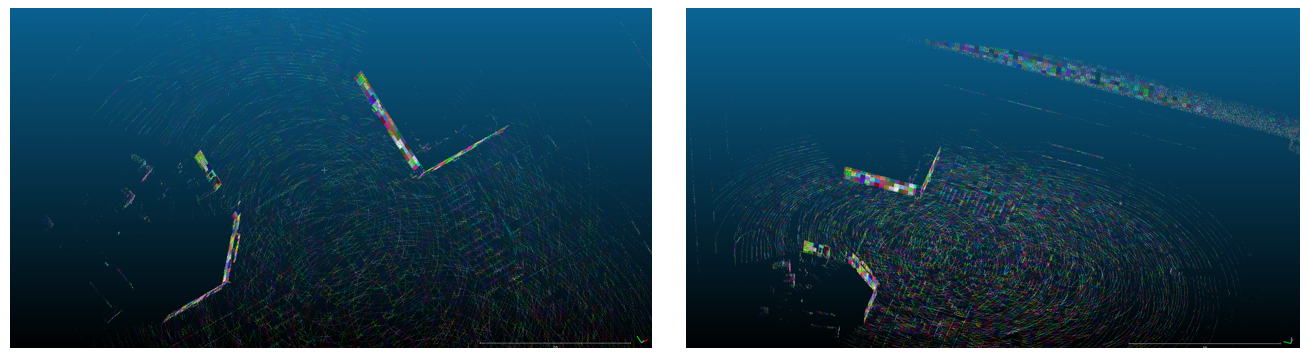} 
\caption{Feature maps extraction for LiDAR to IMU calibration} 
\label{Fig.lidar2imu_feature} 
\end{figure}
It is recommended to collect data according to Fig.\ref{Fig.lidar2imu}. There are several requirements for the calibration scene: a) ensure that the ground is sufficiently flat; b) ensure that there are enough features around, such as walls, lane lines, poles, stationary vehicles, etc; c) the calibrated vehicle circulates three times according to the trajectory shown in Fig.\ref{Fig.lidar2imu}, and the vehicle speed is maintained at 10km/h; d) try not to have dynamic objects around, such as vehicles, etc. Fig.\ref{Fig.lidar2imu_feature} shows feature extraction results for LiDAR to IMU mapping.

\subsubsection{LiDAR to LiDAR calibration}
In autonomous driving, LiDAR plays an important role. LiDAR can obtain the three-dimensional structure information of the surrounding environment in real-time. It can generally build high-precision maps, positioning, obstacle detection, tracking and prediction in autonomous driving. Installing a single LiDAR on a self-driving car sometimes cannot cover the car's surrounding area or does not meet the needs of blind-spot monitoring. Therefore, it is necessary to appropriately increase the number of LiDARs to increase the visible range. The data fusion of multiple LiDARs requires extrinsic parameter calibration of the coordinate systems of multiple LiDARs to obtain each coordinate's accurate rotation and translation parameters for subsequent data processing. We developed a tool for multi-LiDAR calibration based on our previous work \cite{2203.03182}.

The extrinsic calibration between the two LiDARs is achieved in two steps, ground plane alignment and registration of non-ground feature points. First, the point cloud is extracted and segmented from the ground and non-ground point clouds. Then the ground normal is used for ground registration to obtain the roll angle, pitch angle and z-axis translation as initial extrinsic parameters. Then traverse the yaw angle of the transformed non-ground point cloud, and calculate the distance of the closest point of the two LiDARs to obtain the yaw angle of the minimum distance. Subsequently, continue to improve the calibration accuracy through normal iterative closest point (NICP) \cite{serafin2015nicp} and octree-based optimization.


In the rough calibration, we find that the LiDAR can easily sample a large amount of ground plane information on the road. Therefore, the first step of our algorithm is rough registration by taking advantage of the characteristic.
Suppose the maximal plane which contains the most points is considered as the ground plane $GP:\{a,b,c,d\}$: 
\begin{equation}\label{GP extration}
\begin{aligned}
GP:\{a,b,c,d\} = \arg\underset{\lvert plane \rvert}{\max} \lvert {ax_i + by_i + cz_i + d \rvert \leq \epsilon} 
\end{aligned} 
\end{equation}
where $(x_i,y_i,z_i) \in plane$ and $plane \subset  PC$ and $\epsilon$ means the threshold of the plane thickness.
The ground plane is used to align the slave LiDAR ground planes $GP_{s}$ to the master LiDAR ground plane $GP_{m}$:
\begin{equation}\label{GP registration axis}
\begin{aligned}
\vec{n} = \overrightarrow{GP_m} \times \overrightarrow{GP_s}
\end{aligned} 
\end{equation}

\begin{equation}\label{GP registration angle}
\begin{aligned}
\theta = \overrightarrow{GP_m} \cdot \overrightarrow{GP_s}
\end{aligned} 
\end{equation}
where $\vec{n}$, $\theta$, $\overrightarrow{GP_m}$, $\overrightarrow{GP_s}$ represent the rotation axis, rotation angle, master LiDAR normal vector and slave LiDAR normal vector, respectively. The transformation matrix can be computed by Rodriguez formula. It is worth noting that an extreme case can appear where the difference between the estimated pitch/roll and the actual pitch/roll is $\pm \pi$. So the method need to check whether most of the points of $PC_{s}$ are on the ground plane after the calibration. Through above measures, a rough estimation of $angle_{pitch}, angle_{roll}, z $ can be established. The next step is the calibration of $angle_{yaw}, x, y $. The cost function can be simplified as:
\begin{equation}\label{simplified cost function}
\begin{aligned}
&angle^{*}_{yaw} ,\ x^*,\ y^*\ = 
\\ \ &\arg\underset{yaw,x,y}{\min}\sum \sqrt{( x_{s} -x_{m})^{2} +( y_{s} -y_{m})^{2}} 
\end{aligned} 
\end{equation}
The number of arguments decreases from 6 to 3. More importantly, the ground points could be ignored. 



In refinement calibration, we adopt the NICP which is a variant of iterative closest points (ICP) and can achieve better performance than ICP. We suppose that due to the sparsity of the point cloud, the point cloud feature is not explicit and hard to extract. The ICPN enriches the point feature by containing the normal of each point and expands the receptive field of every point.
\begin{figure}[ht]
\centering
\includegraphics[scale=0.4]{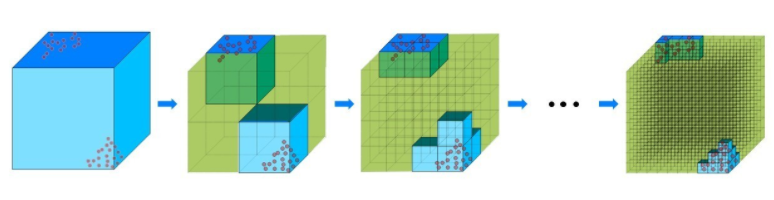}
\caption{Octree-based method. Mark the cube with points in blue and mark the cube without a point in green, cutting the cube iteratively and the volume of blue/green cubes can be measured the quality of calibration.}
\label{octree}
\end{figure}
Furthermore, we continue minimizing the pose error with the octree-based optimization as illustrated in Fig.\ref{octree}. At the beginning, there are two point clouds $PC$ wrapped in a cube $_{o}C$. We then utilize the octree-based method to equally cut the cube into eight smaller cubes. 
\begin{equation}\label{cut the cube}
\begin{aligned}
 _{p}C \stackrel{cut }{\Longrightarrow}  \{_{c}C_1, _{c}C_2,\cdots, _{c}C_7, _{c}C_8\}
\end{aligned} 
\end{equation}
where $_{p}C$ represents the parent cube and $_{c}C_{i}$ represent the child cube of $_{p}C$. The cutting procedure is iteratively repeated to get more and smaller cubes. We mark the cubes with points as blue ones and the cubes without a point as the green ones, as shown in Fig.\ref{octree}. They are further denoted as $C^b$ and $C^g$. The volume of $_{o}C$ can be expressed as follow:
\begin{equation}\label{calculate the space}
\begin{aligned}
V_{_{o}C} = \sum\limits _{i=1}^{N} V_{C^b_i} + \sum\limits _{j=1}^{M} V_{C^g_j}
\end{aligned} 
\end{equation}
where $N$ and $M$ refer to the number of $C^b$ and $C^g$. When the side length of the small cube is short enough, We can approximate that the space volume occupied by the point cloud is the volume of the blue cubes. When two point clouds are aligned accurately, the space volume occupied by point clouds reaches the minimum, and the volume of blue cubes reaches the minimum at the same time. So the problem can be converted to:

\begin{equation}\label{octee-based optimal}
\begin{aligned}
\bm{R}^*,\bm{T}^*\ =\ \arg\underset{\bm{R} ,\bm{T}}{\min} \sum\limits _{j=1}^{M} V_{C^b_j}
\end{aligned} 
\end{equation}
Considering that the current pose is close to the correct, we continue optimizing the formula above by scanning the domain of arguments.

\subsection{Factory Calibration Tools}
Factory calibration is usually the last process in vehicle production. The calibration equipment is mainly composed of multiple calibration boards and four-wheel positioning. After the four-wheel positioning of the vehicle is finished, the calibration starts, and the pose relationship between the sensors is calculated through the calibration board. We provide six types of calibration boards, and the calibration board recognition program removes the OpenCV \cite{opencv_library} library dependency. The recognition performance is very high in different environments. 
\subsubsection{Calibration Board Setup Tools}
\begin{figure}[h]
\centering 
\includegraphics[width=0.5\textwidth]{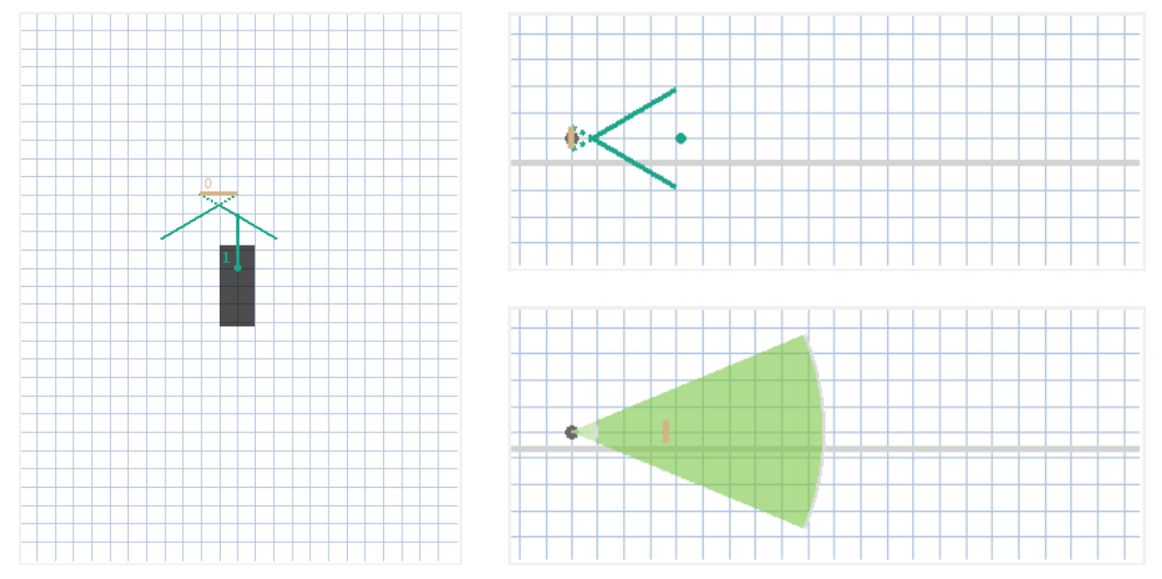} 
\caption{Factory calibration.} 
\label{Fig:factory_setup} 
\end{figure}

Factory calibration requires the construction of a calibration production line, which is usually composed of four-wheel alignment and various calibration boards. 
The pose of the calibration board is related to the position and angle of the vehicle sensor installation, so we developed a tool to automatically generate the optimal position and angle of the calibration board placement. At the same time, the tool can also generate the placement pose range in which the calibration board is valid according to the sensor scheme of the vehicle. 
In addition, if the production line environment has been determined, the tool can also automatically determine whether the sensor scheme of a vehicle is suitable for the current calibration setup. Fig.~\ref{Fig:factory_setup} shows the operation interface of the tool.
\subsubsection{Calibration board detection} This paper mainly introduces the corner detection process of five kinds of calibration boards that can be used on the production line. The five calibration boards are chessboard, circle board, vertical board, aruco marker board, and round hole board. 

\begin{figure}[h]
\centering 
\includegraphics[width=0.5\textwidth]{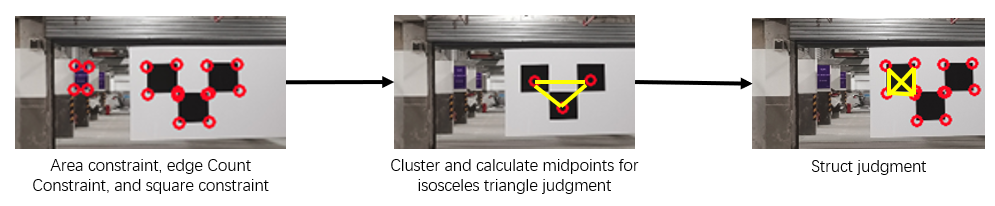} 
\caption{Checkerboard calibration board detection process.} 
\label{Fig.chessboard_flow} 
\end{figure}

\begin{figure}[h]
\centering 
\includegraphics[width=0.45\textwidth]{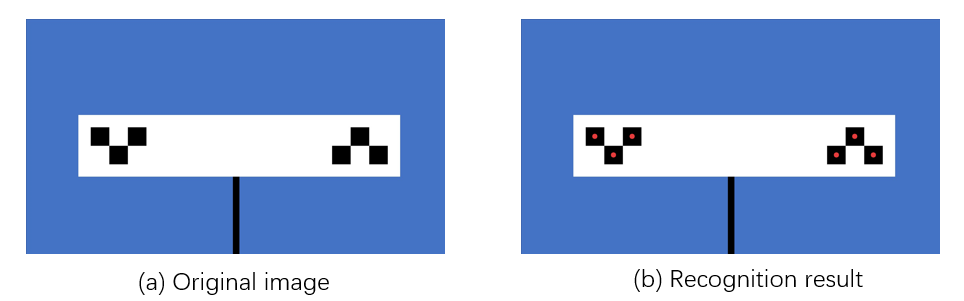} 
\caption{Checkerboard calibration board and detection results.}
\label{Fig.chessboard} 
\end{figure}

For the chessboard (horizontal calibration board) on the production line (generally 2.5 meters wide), the corner detection process first selects the initial threshold to perform adaptive binarization on the image. It then performs the dilate operation to separate the connection of each black block quadrilateral. Because there are no checkerboard corners at the image edge, the edge of the image is set to white. Traverse the image set the black point 0 to white 255 if it is black in the eight areas, and this is to get the edge line of the black block. Then fit the polygon to the black border. Then perform a series of filtering conditions, such as area constraints, rectangle judgment, square judgment to filter out candidate checkerboard regions, cluster the candidate squares to get rough corner points, and finally perform structured judgment on the detected corner points. For example, whether to form an isosceles right triangle with the same area, judge the slope of two points, etc., and output the final corner detection result. Fig.\ref{Fig.chessboard_flow} and Fig.\ref{Fig.chessboard} show show the detection process and detection results of the checkerboard calibration board.

The circular calibration board can be very accurate because all pixels on the periphery of the circle can be used, reducing the effect of image noise.
And in the case of missed detection of multiple corner points, the redundant corner point information can be used for incomplete corner point completion without affecting the calibration process. The calibration process first performs adaptive threshold binarization on the calibration image and then detects the contour of the calibration binary image, extracts circles from the set of image contours, and obtains a set of multiple circles. By judging the distance from the contour pixel to the center of the contour, if the difference between the maximum distance and the minimum distance satisfies the threshold, the contour is considered to be a circle. The circle's center point's color record detects whether the circle is black or white. Line segments are extracted for all circular centers by Ransac, vertical straight lines are extracted, and vertical straight lines are obtained according to the slope range. According to the four constraint filtering conditions, the line segment of the black circle is obtained (1. The two-line segments should be parallel. 2. The corresponding black circle radius is similar. 3. The projection of the corresponding black circle onto the opposite line is coincident. 4. The two-line segments Spacing and black circle radius size constraints.) determine the position of the center of the calibration board. The parallel black lines fix the position of the pattern very well. According to the radius of the black circle, the radius of the white circle and the circle spacing on the line segment are predicted, and finally, the center of all circles on the calibration board is obtained. 
\begin{figure}[h]
\centering 
\includegraphics[width=0.5\textwidth]{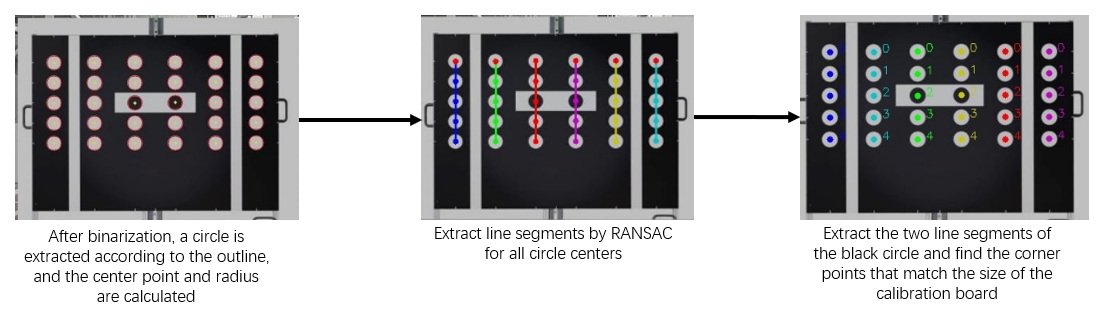} 
\caption{Circle calibration board detection process.} 
\label{Fig.circle_board_flow} 
\end{figure}
\begin{figure}[h]
\centering 
\includegraphics[width=0.45\textwidth]{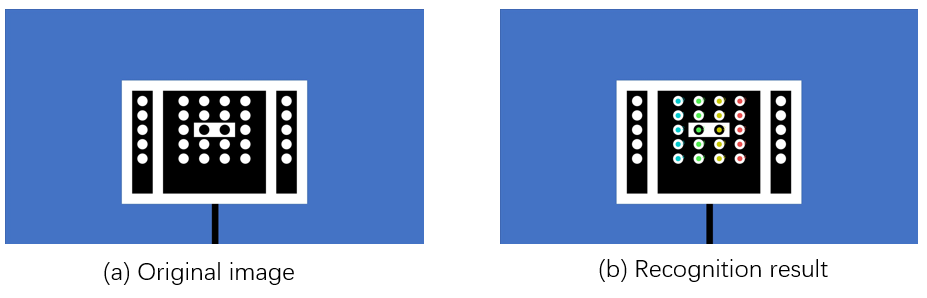} 
\caption{Circle calibration board and detection results.} 
\label{Fig.circle_board} 
\end{figure}

The vertical board is also robust to multiple corner misses. First, the calibration image is gray-scaled, and then all the corners in the image are detected, and a line model, including the corners, is established by randomly selecting points. Each line model is clustered and segmented according to the density of the points. Then find a combination of three parallel and equally spaced lines. For each straight line combination, project the corner points on the left and right straight lines to the middle straight line, and obtain the correct corner points through screening and segmentation through the distance between the corner points on the projected line and the pattern features of the calibration board.
\begin{figure}[h]
\centering 
\includegraphics[width=0.5\textwidth]{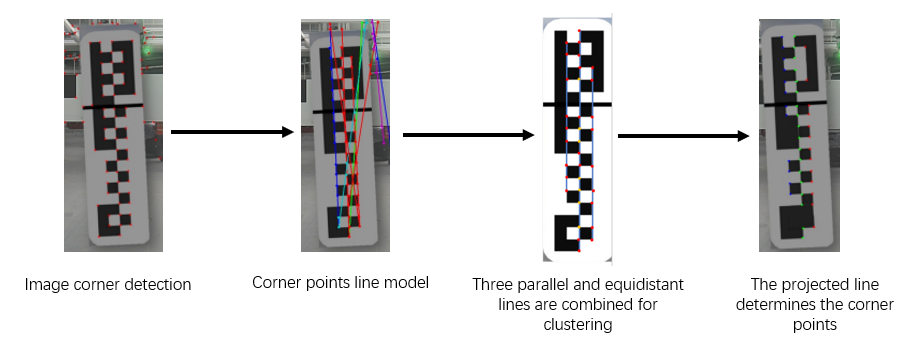} 
\caption{Vertical calibration board detection process.} 
\label{Fig.vertical_board_flow} 
\end{figure}
\begin{figure}[h]
\centering 
\includegraphics[width=0.45\textwidth]{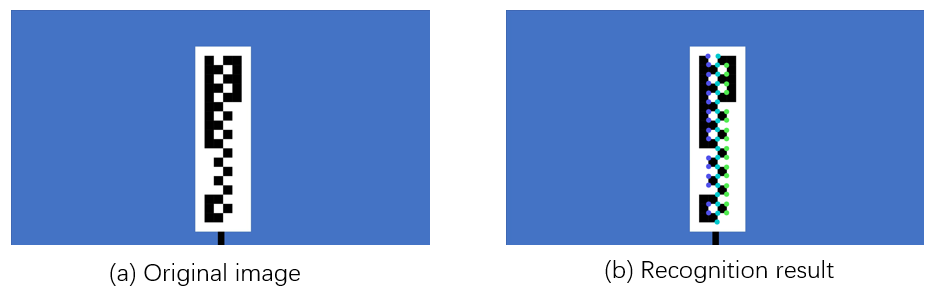} 
\caption{Vertical calibration board and detection results.} 
\label{Fig.vertical_board} 
\end{figure}

The corner detection process of the ArUco marker board first grayscales the RGB image and then uses the method of local adaptive threshold to draw the image binary. Then, a connected domain search is performed to find candidate contours, and contours are screened by the filter condition limited by the number of edges. Then, the polygon fitting based on the contour point set is completed. The fitted polygon conforms to the convex quadrilateral and has a certain distance from the edge of the image, and the contour is not repeated. The sorting of the corner points of the convex quadrilateral after this screening prevents the occurrence of cross-order. Filter the quadrilaterals that are too close. Then, extract the outer rectangle from the quadrilateral through radial transformation, de-binarize the image with a threshold of 127, cut the image to obtain the QR code area, and divide the QR code with a 6*6 grid. Area, encode the corresponding QR code area, match the QR code codebook in the base library, identify the corresponding QR code id, and obtain the corner coordinates on the calibration board.
\begin{figure}[h]
\centering 
\includegraphics[width=0.5\textwidth]{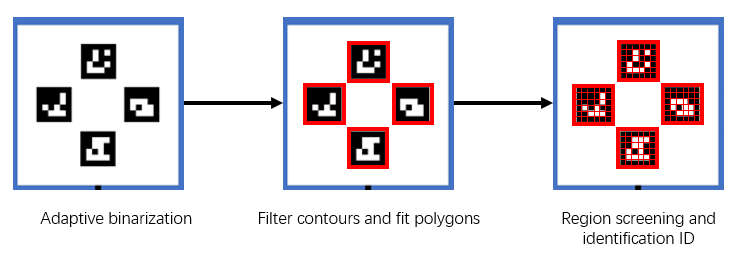} 
\caption{ArUco calibration board detection process.} 
\label{Fig.aruco_marker_board_flow} 
\end{figure}
\begin{figure}[h]
\centering 
\includegraphics[width=0.45\textwidth]{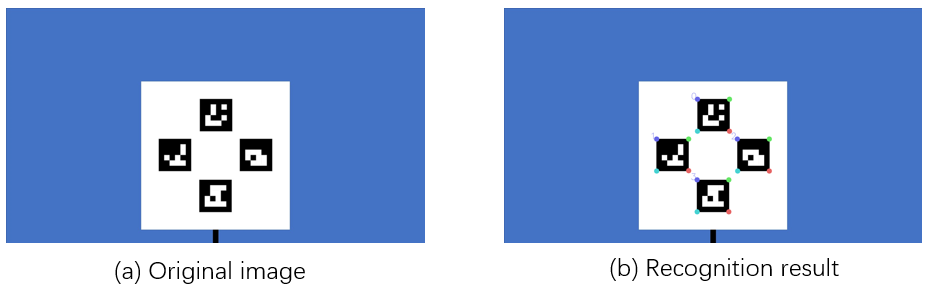} 
\caption{ArUco calibration board and detection results.} 
\label{Fig.aruco_marker_board} 
\end{figure}

The corner detection process of the circular hole board mainly relies on the pattern matching on the calibration board to extract the circle. First, according to the calibration board's size, the circular hole's geometric matching mask is designed. Then carry out a two-dimensional search and matching, and finally obtain the circle and the circle center with the least number of point clouds in the hole.
ROI filtering is performed on the point cloud according to the calibration board's position, and the point cloud near the calibration board is obtained. The calibration board plane is extracted by the RANSAC algorithm with orientation constraints. A calibration board box is preset according to the size of the calibration board, and then the 2D box is extracted and fitted in the point cloud. The initial position of the hole mask is obtained from the size of the mask relative to the box. A mask 2D search is performed near the initial position. Finally, get the coordinates of the center of the circle with the least number of points on the mask.
\begin{figure}[h]
\centering 
\includegraphics[width=0.5\textwidth]{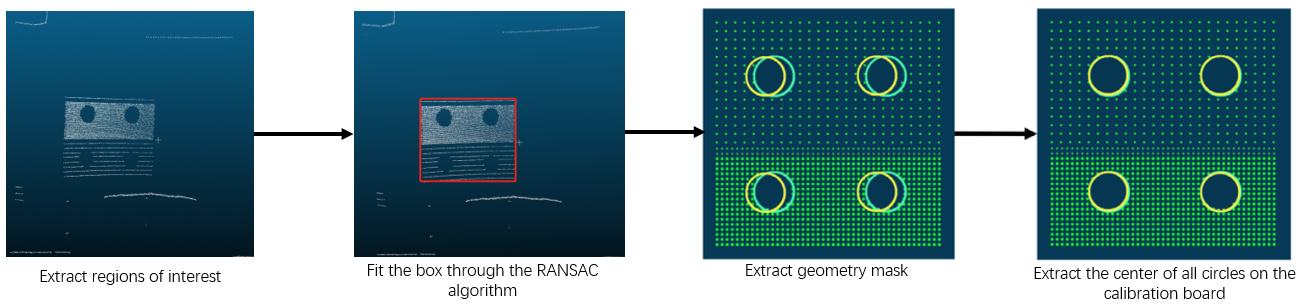} 
\caption{Circular hole calibration board detection process.} 
\label{Fig:round_hole_flow} 
\end{figure}
\begin{figure}[h]
\centering 
\includegraphics[width=0.5\textwidth]{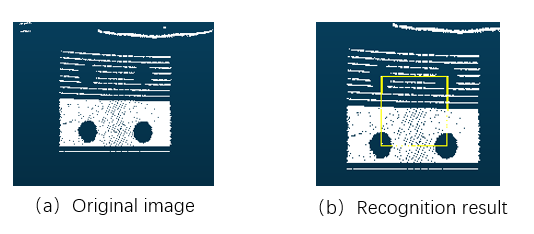} 
\caption{Circular hole calibration board and detection results.} 
\label{Fig:round_hole} 
\end{figure}

AprilTag is a visual reference system that can be used for various tasks, including augmented reality, robotics, and camera calibration. AprilTag detection can calculate the calibration board's exact 3D position, orientation, and id relative to the camera. AprilTag \cite{krogius2019iros} library is implemented in C programming language with no external dependencies. The library can be easily included in other applications or ported to embedded devices. 

\subsubsection{Camera calibration}
\begin{figure}[h]
\centering 
\includegraphics[width=0.48\textwidth]{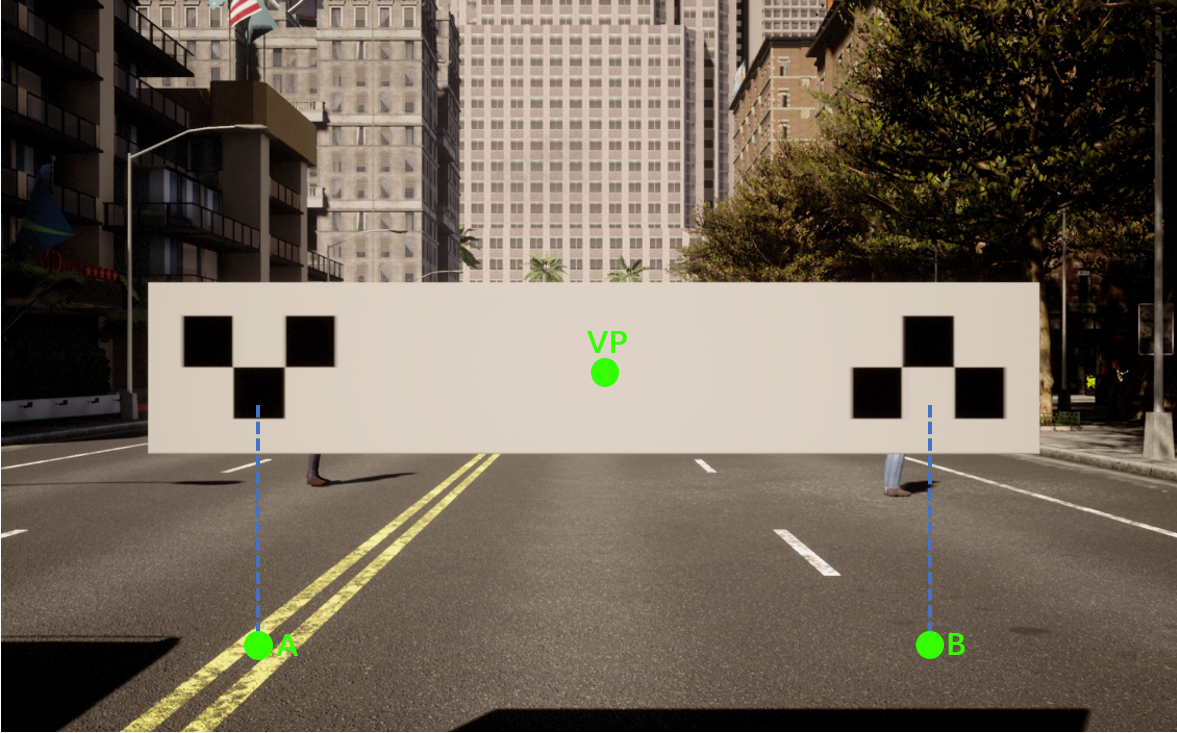} 
\caption{Factory calibration.} 
\label{Fig:factory_calib} 
\end{figure}

The method of batch calibration on the automobile production line is called factory calibration. When the ADAS equipment cannot function normally for other reasons after the vehicle is sold, after-sales calibration of the camera is required. Usually, the camera calibration includes the calibration of the vanishing point of the camera and the homography matrix of the camera to the ground, and the camera-car extrinsic parameter calibration. The calibrated vanishing point is the visual intersection of parallel lines that are aligned with the body of the vehicle. In the factory calibration, the calibration board and the camera positions are fixed. Before the factory calibration, the car to be calibrated needs to be aligned with four wheels. After the four-wheel alignment, the coordinates of the calibration board relative to the center of the vehicle body are fixed, and then the camera is used to identify the calibration board and calibrate. Then, the pose of the camera relative to the vehicle body coordinates can be obtained, which is usually solved by the PnP algorithm \cite{lepetit2009epnp}. The following mainly introduces the calibration method of the vanishing point and the camera-to-ground homography matrix. Because the vanishing point we seek is the intersection of parallel lines parallel to the vehicle body coordinates, then the vanishing point's calculation method is a line of the camera passing through the calibration board, and the line is kept parallel to the vehicle body as shown in Fig.~\ref{Fig:factory_calib}. 
From the homography matrix of the calibration board, and the horizontal and vertical offsets of the camera relative to the center of the calibration board, we can calculate the pixel coordinates of the vanishing point by the following formula.
\begin{equation}
\begin{aligned}
\begin{bmatrix}x \\ y \\ w \end{bmatrix} = 
\begin{bmatrix}
h_{11} & h_{12} & h_{13}\\
h_{21} & h_{22} & h_{23}\\
h_{31} & h_{32} & h_{33}\\
\end{bmatrix}
\begin{bmatrix}X_{offset} \\ Y_{offset} \\ 1 \end{bmatrix}
\\
\end{aligned}
\label{}
\end{equation}

\begin{equation}
\begin{aligned}
    vp_x&=\frac{x}{w},\quad vp_y&=\frac{y}{w}
\end{aligned}
\label{}
\end{equation}
As shown in the figure below, we obtain the homography matrix from the calibration board to the camera according to the actual size of the calibration board and the detected corner points, and then obtain points A and B according to the distance from the calibration board to the ground, as shown in the following figure, Then connect the AB point and the vanishing point, because A-VP and B-VP are parallel lines, so the abscissa is the same, and the ordinate can be obtained by the ranging formula, and then according to the pixel coordinates and world physical coordinates of the 4 points, We can solve the H matrix of the ground relative to the camera.
\begin{figure}[h]
\centering 
\includegraphics[width=0.47\textwidth]{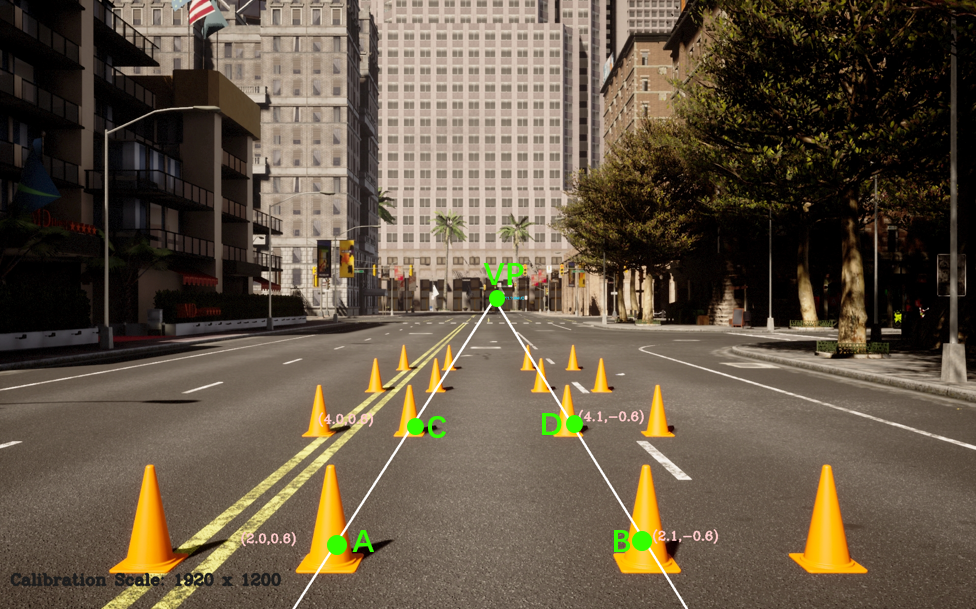} 
\caption{Calibration board to ground homography matrix calibration.} 
\label{Fig.vp} 
\end{figure}
Following~\cite{mobileye}, the camera's ranging results can be obtained through $f_x$ and $f_y$ (camera intrinsic), $H^{\mathrm{C}}$ and ${vp}^{\mathrm{I}}$.
Suppose ($p_x$, $p_y$) is the projected pixel coordinate of a random lane point $\mathbf{p}^\mathrm{L}$. Following perspective principle, the longitudinal and lateral camera ranging for point $\mathbf{p}^\mathrm{L}$ could be defined respectively as follows:
\vspace{-1mm}
\begin{equation}
\footnotesize
\begin{split}
    D_{lon}^\mathrm{C}(\mathbf{p}^\mathrm{L}) = \frac{H^{\mathrm{C}} \cdot f_y}{p_y-{vp}_y^{\mathrm{I}}},  \   
    D_{lat}^\mathrm{C}(\mathbf{p}^\mathrm{L}) = \frac{D_{lon}^\mathrm{C}(\mathbf{p}^\mathrm{L}) ({vp}_x^{\mathrm{I}}-p_x)}{f_x}.
\vspace{-7mm}
\label{Ranging_formula}
\end{split}
\end{equation}
Because we know the actual distance between points A and B, we can optimize the focal lengths $f_x$ and $f_y$ through the above ranging formula, and then calculate the longitudinal coordinates of C and D. Finally, we get the pixel points of the four ground points and the actual physical distance, and then we can calculate the homography matrix from the camera to the ground.
\subsubsection{LiDAR calibration}
As shown in Fig.~\ref{Fig:round_hole}. we detect the three-dimensional coordinates of the center of the LiDAR circle, and at the same time know the position of the calibration board relative to the coordinates of the vehicle body. Therefore, we can obtain the pose of the LiDAR relative to the coordinates of the vehicle body through three sets of non-collinear point pairs. So we constrain the displacement parameters when solving, and then obtain multiple pairs of points through the optimization method.
\begin{equation}
\begin{aligned}
    T_{lidar2car} * P_{lidar}&= P_{car}
\end{aligned}
\label{}
\end{equation}

\begin{equation}
\begin{aligned}
\begin{bmatrix}
\boldsymbol{P_{car}^{1}}&\boldsymbol{P_{car}^{2}} &  \boldsymbol{P_{car}^{3}}
\end{bmatrix} &=
\begin{bmatrix}
\boldsymbol{R} & \boldsymbol{t} 
\end{bmatrix} * 
\begin{bmatrix}
\boldsymbol{P_{lidar}^{1}} & \boldsymbol{P_{lidar}^{2}} &  \boldsymbol{P_{lidar}^{3}} \\
1 & 1 & 1
\end{bmatrix}
\end{aligned}
\label{}
\end{equation}
Because the coordinate position of the lidar relative to the body is relatively accurate, we can add the following constraints.
\begin{equation}
\begin{aligned}
    ||t_1 - a||_2 < \lambda \\
    ||t_2 - a||_2 < \lambda \\
    ||t_3 - a||_2 < \lambda 
\end{aligned}
\label{}
\end{equation}
Among them, $a$, $b$, and $c$ are the installation displacements of the camera in the vehicle body coordinate system, and $\lambda$ is the error value, which is generally a small value. The specific value is set according to the installation tolerance.

\subsubsection{After-sale calibration}
\begin{figure}[h]
\centering 
\includegraphics[width=0.5\textwidth]{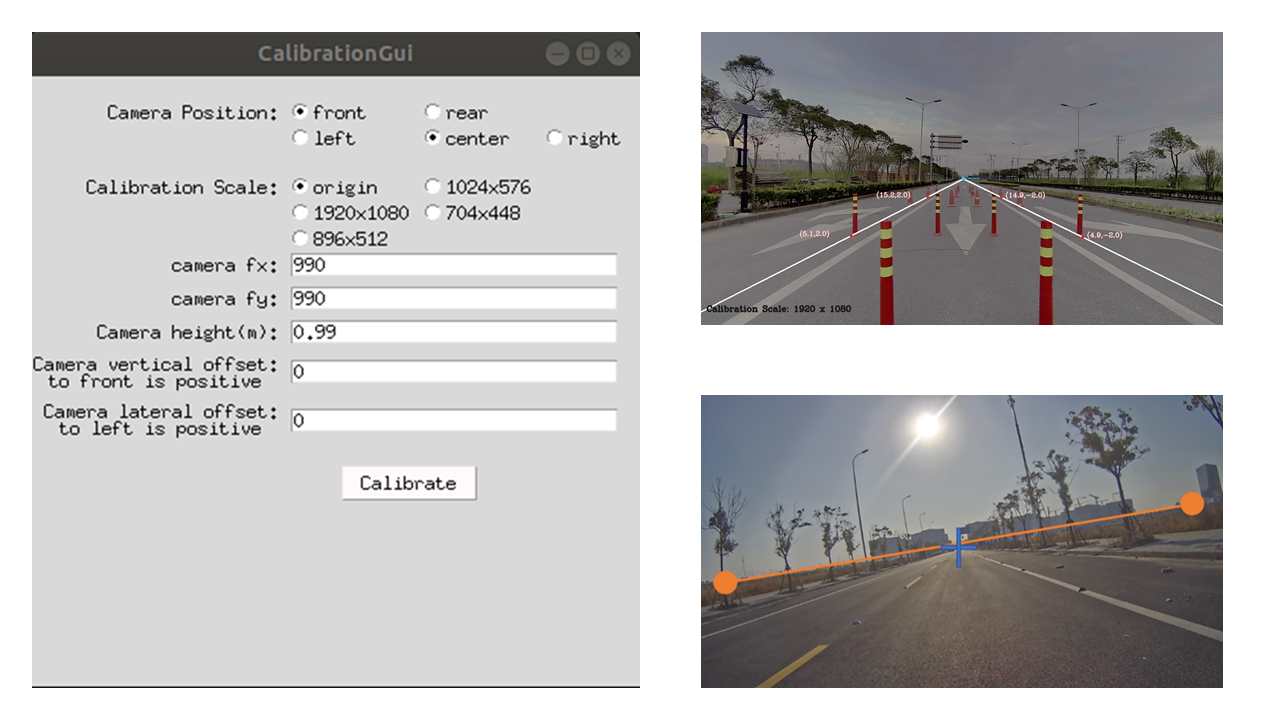} 
\caption{After-sale calibration.} 
\label{Fig:after-sale_calib} 
\end{figure}

After-sales calibration can place the calibration board like the production line calibration, but it is more complicated. Another is to use some environmental features for calibration. Then this section will introduce the use of environmental characteristics for after-sales camera calibration. As shown in the figure, first we need two reference objects parallel to the car body, such as lane lines or other reference objects. Another way is to place 4 reference points to simulate two parallel lines. Then we have 4 points on the parallel line (two on the left and two on the right). According to these four selection points, two straight lines can be fitted, and then the corner points of these two lines are calculated as the vanishing point. The calculation method of the homography matrix can calculate the actual physical coordinates through the camera ranging model such as eq.~\eqref{Ranging_formula}, and then perform the calculation. However, this method has a disadvantage, because the influence of the roll angle will cause an error in the calculation of the lateral physical distance. In order to correct this error, we introduce a roll angle evaluation line. The user drags two points to make the line shorter than the ground. The roll angle can be determined by paralleling the lines with the same vertical distance. We can correct the error of the camera model through the roll angle, and finally calculate the real physical distance of the selected 4 pixels, and then calculate the homography matrix from the camera to the ground. Fig.~\ref{Fig:after-sale_calib} shows the operation interface of our tool and the information that needs to be entered. If you want to verify the accuracy of the calibration, you can compare the actual measurement distance of the selected point with the distance measurement results. If the difference is small, the calibration is considered qualified.
\subsection{Online Calibration Tools}
The traditional calibration method is often in the form of offline, which requires a series of operations with a hand-held calibration board, so it is time-consuming and labor-intensive. The online calibration method refers to the completion of the calibration at the beginning of the system operation or during the operation of the system. This method does not require a hand-held calibration board and can also ensure sufficient accuracy.

Therefore, our calibration toolbox provides users with three online calibration tools for automatic calibration of the vehicle during driving. Likewise, road scenes with strong and easily recognizable features such as trees and traffic signs will lead to better calibration results. The following will introduce the principle of our online calibration.
\subsubsection{Camera to IMU calibration}
Our online calibration between cameras and IMU includes temporal calibration and extrinsic calibration. In the online calibration process, we need a set of image frame data and IMU measurement data of the vehicle during driving. On the first set of frames, we perform temporal calibration, using the trace correlation method \cite{qiu2020real}. Then the temporal calibration between camera and IMU can be represented as:
\vspace{-2mm}
\begin{equation}
\label{equ:temporal calibration}
    \mathbf{td} ^*= \mathbf{argmax}_{\mathbf{td} \in \tau}\mathbf{\Bar{r}}(\Bar{\mathbf{\omega}_I}, \mathbf{\omega}_C)
\vspace{-2mm}
\end{equation}
 
 where $td^*$represents the best time offset estimation between camera and IMU, $\tau$ represents the collection of time offsets, $\Bar{r}(\Bar{\omega_I}, \omega_C)$ represents the trace correlation between $\Bar{\omega_I}$ and $\omega_C$,$\omega_C$ represents camera angular velocity between two frames, and $\Bar{\omega_I}$ represents IMU average angular velocity between two camera frames.
 
 After obtaining the temporal calibration result, we use the result for data alignment.The next step is Rotation calibration,the camera rotation from i frame to i+1 frame is denoted as $\mathbf{RC}_i^\mathrm{i+1}$,the IMU rotation from i frame to i+1 frame is denoted as $\mathbf{RI}_i^\mathrm{i+1}$,the rotation between camera and IMU is denoted as $\mathbf{RIC}$ . Then the rotation calibration can be  represented as:
 \vspace{-2mm}
\begin{equation}
\label{equ:rotation calibration}
    \mathbf{RI}_i^\mathrm{i+1} \cdot \mathbf{RIC} = \mathbf{RIC} \cdot \mathbf{RC}_i^\mathrm{i+1}
\vspace{-2mm}
\end{equation}
 
The final step is translation calibration. we denote camera translation from i frame to i+1 frame as $\mathbf{tc}_i^{i+1}$, IMU translation from i frame to i+1 frame is denoted as  $\mathbf{ti}_i^{i+1}$, while the translation between camera and IMU is denoted as $\mathbf{tic}$. Then the translation calibration can be represented as:
 \vspace{-2mm}
\begin{equation}
\label{equ:translation calibration}
    \mathbf{RC}_i^\mathrm{i+1} \cdot \mathbf{tic} - \mathbf{tic} = \mathbf{RIC} \cdot \mathbf{ti}_i^{i+1} - \mathbf{tc}_i^{i+1}
\vspace{-2mm}
\end{equation}
\subsubsection{Lidar to IMU calibration}
Lidar to IMU online calibration is similar to Camera to IMU calibration, We used the front-end of LOAM as Lidar odometry, then align lidar data with IMU data, finally, we can use a similar method to get results of Lidar to IMU online calibration. 
the Lidar rotation from i frame to i+1 frame is denoted as $\mathbf{RL}_i^\mathrm{i+1}$,the IMU rotation from i frame to i+1 frame is denoted as $\mathbf{RI}_i^\mathrm{i+1}$,the rotation between camera and IMU is denoted as $\mathbf{RIL}$ . Then the rotation calibration can be  represented as:
 \vspace{-2mm}
\begin{equation}
\label{equ:rotationL2I calibration}
    \mathbf{RI}_i^\mathrm{i+1} \cdot \mathbf{RIL} = \mathbf{RIL} \cdot \mathbf{RL}_i^\mathrm{i+1}
\vspace{-2mm}
\end{equation}
 
The final step is translation calibration. we denote camera translation from i frame to i+1 frame as $\mathbf{tl}_i^{i+1}$, IMU translation from i frame to i+1 frame is denoted as  $\mathbf{ti}_i^{i+1}$, while the translation between camera and IMU is denoted as $\mathbf{til}$. Then the translation calibration can be represented as:
 \vspace{-2mm}
\begin{equation}
\label{equ:translationL2I calibration}
    \mathbf{RL}_i^\mathrm{i+1} \cdot \mathbf{til} - \mathbf{til} = \mathbf{RIL} \cdot \mathbf{ti}_i^{i+1} - \mathbf{tl}_i^{i+1}
\vspace{-2mm}
\end{equation}

\subsubsection{Radar to Carcenter calibration}In this task, we mainly calibrate the yaw angle of radar. The problem is divided into three steps, rough calibration, static object recognition, and curve fitting. In the first step. We denoted the Radar measurement data as $\{\mathbf{V}_i,\mathbf{angle}_i,\mathbf{distance}_i\}$, where $\mathbf{V}_i$ means the velocity of the target relative to the radial direction of the radar measured by the Doppler principle, $\mathbf{{angle}_i}$ means the angle between the object and the radar coordinate system, and $\mathbf{{distance}_i}$ means the distance between object and car.

our rough calibration by finding the object directly in front of the vehicle, then the radar measurement of $\mathbf{angle_i}$ 
is the result of yaw. The next step is to detect static objects, for a static object satisfying the following equation.
 \vspace{-2mm}
\begin{equation}
\label{equ:static detection}
    \mathbf{V}_i = \mathbf{V}_e \cdot \mathbf{\cos({{angle}_i+yaw})}
\vspace{-2mm}
\end{equation}
where $\mathbf{V}_e$ means vehicle velocity. After finding static objects for all frames, the final step is to fit the cos function curve, which we can get more accurate calibration results.

\subsection{Calibration Benchmark Datasets}
Based on a large number of evaluation experiments and research carried out for the current mainstream models of LiDAR, camera, and other sensors, we have established a sensor feature set, including the detection performance and feature expression of different technical routes and models of autopilot sensors. We have completed the development of the sensor feature set to the sensor simulation library based on the Carla platform \cite{Dosovitskiy17}, an open-source simulation framework for autonomous driving. For example, for LiDAR, according to the results of evaluation experiments, we simulate the detection performance, timing characteristics, and harness distribution of sensors in different scenes according to the rules summarized from the actual data. In addition, weather conditions will also have various point losses and noise effects on different LiDAR models, which will also be reflected in the point cloud data generated by simulation.

Based on the established sensor simulation library, we complete multi-sensor data simulation in different calibration scenarios and finally generate calibration benchmark datasets. In the simulation framework, we control the ego vehicle to complete driving along a specific trajectory in the environment with obvious peripheral references such as lane line, street lamp, wall, and car and record the data required for specific calibration at the same time. The data generation frequency is also different for different sensors, as shown in the table below. To ensure the normal recording of data, the system frequency of the simulation platform should be higher than the recording frequency of any sensor. The multi-calibration scene simulation diagram is shown in the figure below. In this figure, (a) is the camera intrinsic parameter calibration simulation scene, figure (b) is the online calibration simulation scene, and figure (c) is the Radar extrinsic parameter calibration simulation scene.
\begin{table}[htbp]
\caption{Sensor Data Recording Frequency}
\centering
\renewcommand{\arraystretch}{1.3}
\begin{tabular}{|c|c|c|c|c|c}
\hline
 & IMU & LiDAR & Camera & Radar \\
\hline
FPS & $100$ Hz & $10$ Hz & $10$ Hz & $20$ Hz \\
\hline
\end{tabular}  
\label{table:reprojection} 
\end{table}

\begin{figure}[h]
\centering 
\includegraphics[width=0.5\textwidth]{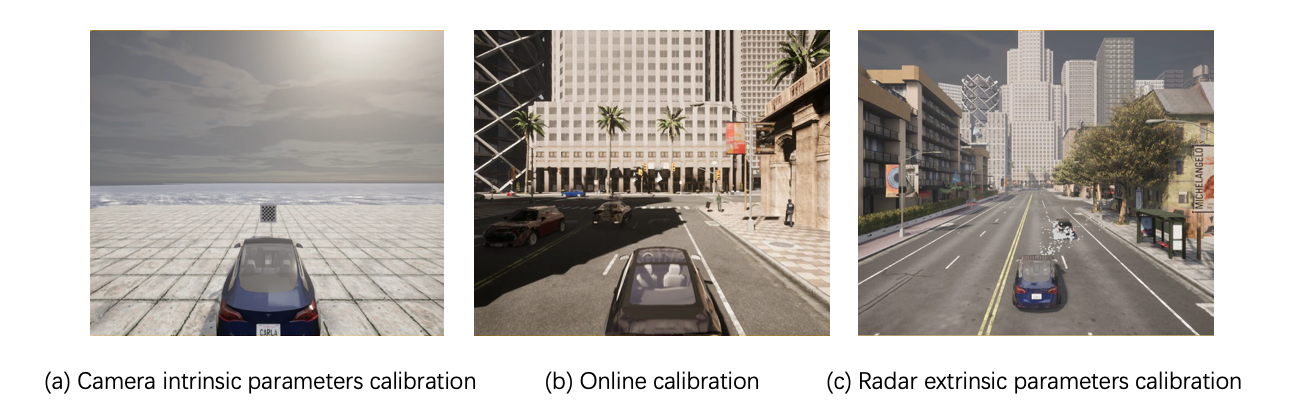} 
\caption{sensors calibration simulation scenes } 
\label{Fig.calibration_simulation} 
\end{figure}

The advantage of simulation data set is that it can obtain multi-sensor data of complex scenes at very low cost, and can provide high-precision truth value to verify the accuracy of calibration algorithm. We have generated a small amount of data for testing the calibration performance in the simulator. In the future, we will generate a large amount of calibration test data in the simulator to form calibration benchmark datasets.

\section{EXPERIMENTS}

\subsection{Experiment Settings}
We conducted experiments on real driverless platforms. The sensors used include Hesai Pandar64 LiDAR, Balser acA1920-40gc Camera with  different FOVs ($FOV=30^{\circ}$, $FOV=60^{\circ}$, $FOV=120^{\circ}$), Delphi ESR 2.5 Radar and NovAtel PP7D. To better qualitatively evaluate the calibration accuracy, we also conducted experiments with our method in the simulation environment. All simulation calibration data are generated by the Carla engine \cite{Dosovitskiy17}, which is introduced in the calibration benchmark datasets in the previous section.

\subsection{Quantitative Results}
Such as manual calibration tools, adjust the extrinsic parameters of the calibration according to the visualization results. On the same basis, to judge the calibration performance and whether the calibration parameters are available, we can fuse different sensors and judge whether the calibration is accurate according to the visualization effect. This method is more intuitive, and the sensor fusion effect is evaluated manually to avoid wrong calibration parameters. The following will introduce the effect of the visual judgment of some calibration tool.

\begin{figure}[h]
\centering 
\includegraphics[width=0.5\textwidth]{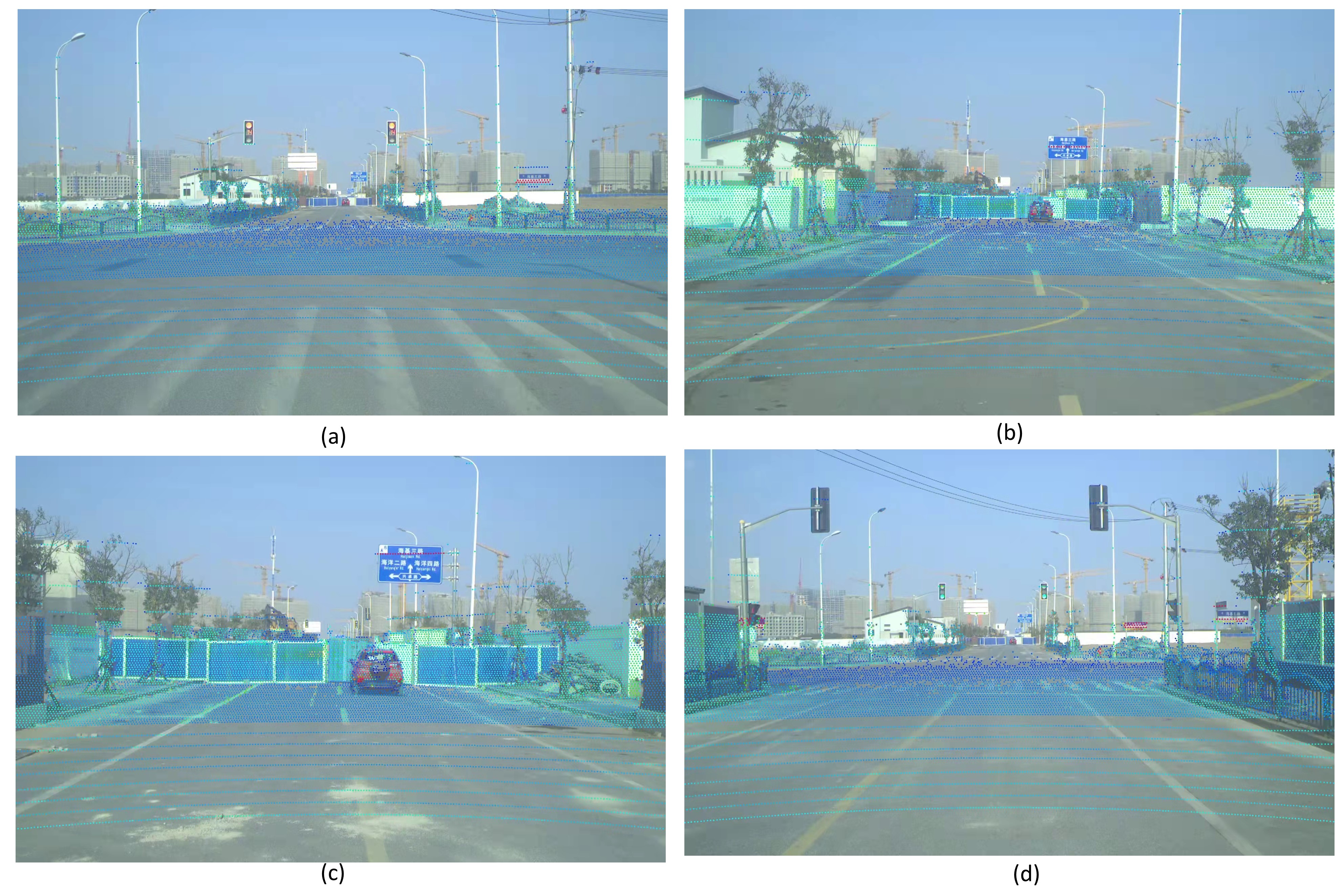} 
\caption{the point cloud is projected to the image plane } 
\label{Fig.reprojection1} 
\end{figure}
\begin{figure}[h]
\centering 
\includegraphics[width=0.5\textwidth]{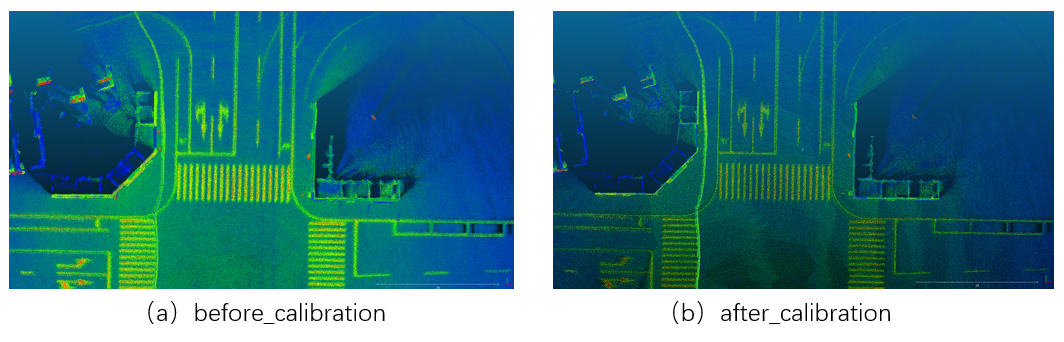} 
\caption{Comparison of point cloud maps before and after calibration} 
\label{Fig.vs_l2i} 
\end{figure}
\begin{figure}[h]
\centering 
\includegraphics[width=0.5\textwidth]{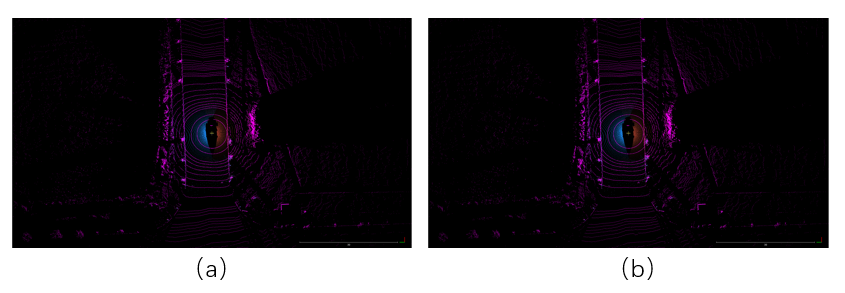} 
\caption{Point cloud map after refine} 
\label{Fig.l2l} 
\end{figure}
To better visualize the performance of LiDAR to the camera calibration, the point cloud is projected to the image plane using the calibration parameter. Fig.~\ref{Fig.reprojection1} shows the projection result of LiDAR point cloud on the image. The more accurate the calibration parameters, the more aligned the object pairs of the point cloud and the image in the projection image.
To better visualize the performance of LiDAR to IMU calibration, 
Multi-frame point clouds are spliced by IMU pose and LiDAR-IMU calibration parameters in Fig. \ref{Fig.vs_l2i}. The more accurate the calibration parameters, the better the mapping effect.
To better visualize the performance of LiDAR to LiDAR calibration, 
Multi-frame point clouds are spliced by LiDAR-LiDAR calibration parameters in Fig. \ref{Fig.l2l}. The higher the calibration accuracy, the more aligned the same object pair is under different LiDARs. Other calibration tools are similar and are not shown here.

\subsection{Qualitative Results}
Quantitative evaluation requires the ground truth, and the simulation environment can provide a very accurate ground truth. In addition, the consistency verification of multiple calibrations in different scenarios can judge the robustness of the calibration method. Therefore, on the real data lacking ground truth, we can use consistency to evaluate the calibration algorithm's robustness and the calibration parameters' accuracy. In the future, we will generate a large amount of calibration data through the simulator. We will use these simulation data to evaluate the calibration accuracy of our calibration algorithm, and we will also open-source the data.

\section{CONCLUSIONS}
Sensor calibration is the foundation block of an autonomous system and its constituent sensors. It must be performed correctly before sensor fusion is implemented. Precise calibrations are vital for further processing steps, such as sensor fusion and implementation of algorithms for obstacle detection, localization and mapping, and control. Further, sensor fusion is one of the essential tasks in autonomous driving applications that fuses information obtained from multiple sensors to reduce the uncertainties compared to when sensors are used individually. To solve the problem of sensor calibration for autonomous vehicles, we provide a sensors calibration toolbox OpenCalib. The calibration toolbox can calibrate sensors such as IMU, LiDAR, Camera, and Radar. This paper mainly introduces the methods of some calibration tools in OpenCalib. If you want to know the specific details, you can check our open-source code. In addition, our toolbox also has some shortcomings to be optimized, and we will continue to optimize our calibration toolbox in the future.

\section{ACKNOWLEDGEMENTS}
We thank the reviewers for their valuable comments. 
The toolbox started from a codebase of the TEDS team of SenseTime Intelligent Driving Group. Thanks to these team members for their contributions to this toolbox.
This work was supported by Shanghai AI Laboratory.

\bibliographystyle{IEEEtran}
\bibliography{egbib}

\begin{thebibliography}{10}
\providecommand{\url}[1]{#1}
\csname url@rmstyle\endcsname
\providecommand{\newblock}{\relax}
\providecommand{\bibinfo}[2]{#2}
\providecommand\BIBentrySTDinterwordspacing{\spaceskip=0pt\relax}
\providecommand\BIBentryALTinterwordstretchfactor{4}
\providecommand\BIBentryALTinterwordspacing{\spaceskip=\fontdimen2\font plus
\BIBentryALTinterwordstretchfactor\fontdimen3\font minus
  \fontdimen4\font\relax}
\providecommand\BIBforeignlanguage[2]{{%
\expandafter\ifx\csname l@#1\endcsname\relax
\typeout{** WARNING: IEEEtran.bst: No hyphenation pattern has been}%
\typeout{** loaded for the language `#1'. Using the pattern for}%
\typeout{** the default language instead.}%
\else
\language=\csname l@#1\endcsname
\fi
#2}}

\bibitem{kato2018autoware}
S.~Kato, S.~Tokunaga, Y.~Maruyama, S.~Maeda, M.~Hirabayashi, Y.~Kitsukawa,
  A.~Monrroy, T.~Ando, Y.~Fujii, and T.~Azumi, ``Autoware on board: Enabling
  autonomous vehicles with embedded systems,'' in \emph{2018 ACM/IEEE 9th
  International Conference on Cyber-Physical Systems (ICCPS)}.\hskip 1em plus
  0.5em minus 0.4em\relax IEEE, 2018, pp. 287--296.

\bibitem{opencv_library}
G.~Bradski, ``{The OpenCV Library},'' \emph{Dr. Dobb's Journal of Software
  Tools}, 2000.

\bibitem{Dosovitskiy17}
A.~Dosovitskiy, G.~Ros, F.~Codevilla, A.~Lopez, and V.~Koltun, ``{CARLA}: {An}
  open urban driving simulator,'' in \emph{Proceedings of the 1st Annual
  Conference on Robot Learning}, 2017, pp. 1--16.

\bibitem{zhang2000}
Z.~Zhang, ``A flexible new technique for camera calibration,'' \emph{IEEE
  Transactions on Pattern Analysis and Machine Intelligence}, vol.~22, no.~11,
  pp. 1330--1334, 2000.

\bibitem{heikkila1997four}
J.~Heikkila and O.~Silv{\'e}n, ``A four-step camera calibration procedure with
  implicit image correction,'' in \emph{Proceedings of IEEE computer society
  conference on computer vision and pattern recognition}.\hskip 1em plus 0.5em
  minus 0.4em\relax IEEE, 1997, pp. 1106--1112.

\bibitem{atanacio2011lidar}
G.~Atanacio-Jim{\'e}nez, J.-J. Gonz{\'a}lez-Barbosa, J.~B. Hurtado-Ramos, F.~J.
  Ornelas-Rodr{\'\i}guez, H.~Jim{\'e}nez-Hern{\'a}ndez, T.~Garc{\'\i}a-Ramirez,
  and R.~Gonz{\'a}lez-Barbosa, ``Lidar velodyne hdl-64e calibration using
  pattern planes,'' \emph{International Journal of Advanced Robotic Systems},
  vol.~8, no.~5, p.~59, 2011.

\bibitem{muhammad2010calibration}
N.~Muhammad and S.~Lacroix, ``Calibration of a rotating multi-beam lidar,'' in
  \emph{2010 IEEE/RSJ International Conference on Intelligent Robots and
  Systems}.\hskip 1em plus 0.5em minus 0.4em\relax IEEE, 2010, pp. 5648--5653.

\bibitem{zhang2004}
Q.~Zhang and R.~Pless, ``Extrinsic calibration of a camera and laser range
  finder (improves camera calibration),'' in \emph{2004 IEEE/RSJ International
  Conference on Intelligent Robots and Systems (IROS) (IEEE Cat.
  No.04CH37566)}, vol.~3, 2004, pp. 2301--2306 vol.3.

\bibitem{2202.13708}
G.~Yan, F.~He, C.~Shi, X.~Cai, and Y.~Li, ``Joint camera intrinsic and
  lidar-camera extrinsic calibration,'' 2022.

\bibitem{wang2011integrating}
T.~Wang, N.~Zheng, J.~Xin, and Z.~Ma, ``Integrating millimeter wave radar with
  a monocular vision sensor for on-road obstacle detection applications,''
  \emph{Sensors}, vol.~11, no.~9, pp. 8992--9008, 2011.

\bibitem{pervsic2017extrinsic}
J.~Per{\v{s}}i{\'c}, I.~Markovi{\'c}, and I.~Petrovi{\'c}, ``Extrinsic 6dof
  calibration of 3d lidar and radar,'' in \emph{2017 European Conference on
  Mobile Robots (ECMR)}.\hskip 1em plus 0.5em minus 0.4em\relax IEEE, 2017, pp.
  1--6.

\bibitem{levinson2013}
J.~Levinson and S.~Thrun, ``Automatic online calibration of cameras and
  lasers.'' in \emph{Robotics: Science and Systems}, vol.~2, 2013, p.~7.

\bibitem{pandey2012}
G.~Pandey, J.~R. McBride, S.~Savarese, and R.~M. Eustice, ``Automatic
  targetless extrinsic calibration of a 3d lidar and camera by maximizing
  mutual information.'' in \emph{AAAI}, 2012.

\bibitem{zhu2020online}
Y.~Zhu, C.~Li, and Y.~Zhang, ``Online camera-lidar calibration with sensor
  semantic information,'' in \emph{2020 IEEE International Conference on
  Robotics and Automation (ICRA)}, 2020, pp. 4970--4976.

\bibitem{ma2021crlf}
T.~Ma, Z.~Liu, G.~Yan, and Y.~Li, ``Crlf: Automatic calibration and refinement
  based on line feature for lidar and camera in road scenes,'' 2021.

\bibitem{barazzetti2011targetless}
L.~Barazzetti, L.~Mussio, F.~Remondino, and M.~Scaioni, ``Targetless camera
  calibration,'' \emph{International Archives of the Photogrammetry, Remote
  Sensing and Spatial Information Sciences}, vol.~38, no. 5/W16, p.~8, 2011.

\bibitem{2203.03182}
P.~Wei, G.~Yan, Y.~Li, K.~Fang, W.~Liu, X.~Cai, and J.~Yang, ``Croon: Automatic
  multi-lidar calibration and refinement method in road scene,'' 2022.

\bibitem{gong20133d}
X.~Gong, Y.~Lin, and J.~Liu, ``3d lidar-camera extrinsic calibration using an
  arbitrary trihedron,'' \emph{Sensors}, vol.~13, no.~2, pp. 1902--1918, 2013.

\bibitem{pandey2015automatic}
G.~Pandey, J.~R. McBride, S.~Savarese, and R.~M. Eustice, ``Automatic extrinsic
  calibration of vision and lidar by maximizing mutual information,''
  \emph{Journal of Field Robotics}, vol.~32, no.~5, pp. 696--722, 2015.

\bibitem{Taylor2012AutomaticCO}
Z.~Taylor and J.~I. Nieto, ``Automatic calibration of lidar and camera images
  using normalized mutual information,'' 2012.

\bibitem{strobl2006optimal}
K.~H. Strobl and G.~Hirzinger, ``Optimal hand-eye calibration,'' in \emph{2006
  IEEE/RSJ international conference on intelligent robots and systems}.\hskip
  1em plus 0.5em minus 0.4em\relax IEEE, 2006, pp. 4647--4653.

\bibitem{shiu1987calibration}
Y.~C. Shiu and S.~Ahmad, ``Calibration of wrist-mounted robotic sensors by
  solving homogeneous transform equations of the form ax= xb,'' 1987.

\bibitem{park1994robot}
F.~C. Park and B.~J. Martin, ``Robot sensor calibration: solving ax= xb on the
  euclidean group,'' \emph{IEEE Transactions on Robotics and Automation},
  vol.~10, no.~5, pp. 717--721, 1994.

\bibitem{daniilidis1999hand}
K.~Daniilidis, ``Hand-eye calibration using dual quaternions,'' \emph{The
  International Journal of Robotics Research}, vol.~18, no.~3, pp. 286--298,
  1999.

\bibitem{fassi2005hand}
I.~Fassi and G.~Legnani, ``Hand to sensor calibration: A geometrical
  interpretation of the matrix equation ax= xb,'' \emph{Journal of Robotic
  Systems}, vol.~22, no.~9, pp. 497--506, 2005.

\bibitem{horn2021online}
M.~Horn, T.~Wodtko, M.~Buchholz, and K.~Dietmayer, ``Online extrinsic
  calibration based on per-sensor ego-motion using dual quaternions,''
  \emph{IEEE Robotics and Automation Letters}, vol.~6, no.~2, pp. 982--989,
  2021.

\bibitem{dornaika1998simultaneous}
F.~Dornaika and R.~Horaud, ``Simultaneous robot-world and hand-eye
  calibration,'' \emph{IEEE transactions on Robotics and Automation}, vol.~14,
  no.~4, pp. 617--622, 1998.

\bibitem{huang2017extrinsic}
K.~Huang and C.~Stachniss, ``Extrinsic multi-sensor calibration for mobile
  robots using the gauss-helmert model,'' in \emph{2017 IEEE/RSJ International
  Conference on Intelligent Robots and Systems (IROS)}.\hskip 1em plus 0.5em
  minus 0.4em\relax IEEE, 2017, pp. 1490--1496.

\bibitem{rehder2016general}
J.~Rehder, R.~Siegwart, and P.~Furgale, ``A general approach to spatiotemporal
  calibration in multisensor systems,'' \emph{IEEE Transactions on Robotics},
  vol.~32, no.~2, pp. 383--398, 2016.

\bibitem{furgale2013unified}
P.~Furgale, J.~Rehder, and R.~Siegwart, ``Unified temporal and spatial
  calibration for multi-sensor systems,'' in \emph{2013 IEEE/RSJ International
  Conference on Intelligent Robots and Systems}.\hskip 1em plus 0.5em minus
  0.4em\relax IEEE, 2013, pp. 1280--1286.

\bibitem{qin2018online}
T.~Qin and S.~Shen, ``Online temporal calibration for monocular visual-inertial
  systems,'' in \emph{2018 IEEE/RSJ International Conference on Intelligent
  Robots and Systems (IROS)}.\hskip 1em plus 0.5em minus 0.4em\relax IEEE,
  2018, pp. 3662--3669.

\bibitem{qiu2020real}
K.~Qiu, T.~Qin, J.~Pan, S.~Liu, and S.~Shen, ``Real-time temporal and
  rotational calibration of heterogeneous sensors using motion correlation
  analysis,'' \emph{IEEE Transactions on Robotics}, vol.~37, no.~2, pp.
  587--602, 2020.

\bibitem{DBLP:journals/corr/SchneiderPSF17}
\BIBentryALTinterwordspacing
N.~Schneider, F.~Piewak, C.~Stiller, and U.~Franke, ``Regnet: Multimodal sensor
  registration using deep neural networks,'' \emph{CoRR}, vol. abs/1707.03167,
  2017. [Online]. Available: \url{http://arxiv.org/abs/1707.03167}
\BIBentrySTDinterwordspacing

\bibitem{2018calibnet}
\BIBentryALTinterwordspacing
G.~Iyer, R.~K. Ram, J.~K. Murthy, and K.~M. Krishna, ``Calibnet: Geometrically
  supervised extrinsic calibration using 3d spatial transformer networks,''
  \emph{2018 IEEE/RSJ International Conference on Intelligent Robots and
  Systems (IROS)}, Oct 2018. [Online]. Available:
  \url{http://dx.doi.org/10.1109/IROS.2018.8593693}
\BIBentrySTDinterwordspacing

\bibitem{RGGnet2020}
K.~Yuan, Z.~Guo, and Z.~J. Wang, ``Rggnet: Tolerance aware lidar-camera online
  calibration with geometric deep learning and generative model,'' \emph{IEEE
  Robotics and Automation Letters}, vol.~5, no.~4, pp. 6956--6963, 2020.

\bibitem{bogdan2018deepcalib}
\BIBentryALTinterwordspacing
O.~Bogdan, V.~Eckstein, F.~Rameau, and J.-C. Bazin, ``Deepcalib: A deep
  learning approach for automatic intrinsic calibration of wide field-of-view
  cameras,'' in \emph{Proceedings of the 15th ACM SIGGRAPH European Conference
  on Visual Media Production}, ser. CVMP '18.\hskip 1em plus 0.5em minus
  0.4em\relax New York, NY, USA: Association for Computing Machinery, 2018.
  [Online]. Available: \url{https://doi.org/10.1145/3278471.3278479}
\BIBentrySTDinterwordspacing

\bibitem{deepvp2018}
C.-K. Chang, J.~Zhao, and L.~Itti, ``Deepvp: Deep learning for vanishing point
  detection on 1 million street view images,'' in \emph{2018 IEEE International
  Conference on Robotics and Automation (ICRA)}, 2018, pp. 4496--4503.

\bibitem{MonoEF2021}
\BIBentryALTinterwordspacing
Y.~Zhou, Y.~He, H.~Zhu, C.~Wang, H.~Li, and Q.~Jiang, ``Monocular 3d object
  detection: An extrinsic parameter free approach,'' \emph{CoRR}, vol.
  abs/2106.15796, 2021. [Online]. Available:
  \url{https://arxiv.org/abs/2106.15796}
\BIBentrySTDinterwordspacing

\bibitem{2017demon}
\BIBentryALTinterwordspacing
B.~Ummenhofer, H.~Zhou, J.~Uhrig, N.~Mayer, E.~Ilg, A.~Dosovitskiy, and
  T.~Brox, ``Demon: Depth and motion network for learning monocular stereo,''
  \emph{2017 IEEE Conference on Computer Vision and Pattern Recognition
  (CVPR)}, Jul 2017. [Online]. Available:
  \url{http://dx.doi.org/10.1109/CVPR.2017.596}
\BIBentrySTDinterwordspacing

\bibitem{kendall2016posenet}
A.~Kendall, M.~Grimes, and R.~Cipolla, ``Posenet: A convolutional network for
  real-time 6-dof camera relocalization,'' 2016.

\bibitem{teed2020deepv2d}
Z.~Teed and J.~Deng, ``Deepv2d: Video to depth with differentiable structure
  from motion,'' 2020.

\bibitem{GeoNet2018}
\BIBentryALTinterwordspacing
Z.~Yin and J.~Shi, ``Geonet: Unsupervised learning of dense depth, optical flow
  and camera pose,'' \emph{CoRR}, vol. abs/1803.02276, 2018. [Online].
  Available: \url{http://arxiv.org/abs/1803.02276}
\BIBentrySTDinterwordspacing

\bibitem{zhou2017unsupervised}
T.~Zhou, M.~Brown, N.~Snavely, and D.~G. Lowe, ``Unsupervised learning of depth
  and ego-motion from video,'' 2017.

\bibitem{stu2006img}
R.~S. Sturm~P. and L.~S, ``Imaging beyond the pinhole camera,'' in \emph{On
  Calibration, Structure from Motion and Multi-View Geometry for Generic Camera
  Models}, 2006, pp. 1--8.

\bibitem{juarez2020distorted}
R.~Juarez-Salazar, J.~Zheng, and V.~H. Diaz-Ramirez, ``Distorted pinhole camera
  modeling and calibration,'' \emph{Applied Optics}, vol.~59, no.~36, pp.
  11\,310--11\,318, 2020.

\bibitem{tang2012high}
Z.~Tang, R.~G. von Gioi, P.~Monasse, and J.-M. Morel, ``High-precision camera
  distortion measurements with a “calibration harp”,'' \emph{JOSA A},
  vol.~29, no.~10, pp. 2134--2143, 2012.

\bibitem{canny1986computational}
J.~Canny, ``A computational approach to edge detection,'' \emph{IEEE
  Transactions on pattern analysis and machine intelligence}, no.~6, pp.
  679--698, 1986.

\bibitem{yu2018bisenet}
C.~Yu, J.~Wang, C.~Peng, C.~Gao, G.~Yu, and N.~Sang, ``Bisenet: Bilateral
  segmentation network for real-time semantic segmentation,'' in \emph{2018
  European conference on computer vision (ECCV)}, 2018, pp. 325--341.

\bibitem{BALM2021}
.~Z.~F. Liu~Zheng, ``Balm: Bundle adjustment for lidar mapping,'' \emph{IEEE
  Robotics and Automation Letters}, vol.~6, no.~2, p. 3184–3191, 2021.

\bibitem{serafin2015nicp}
J.~Serafin and G.~Grisetti, ``Nicp: Dense normal based point cloud
  registration,'' in \emph{2015 IEEE/RSJ International Conference on
  Intelligent Robots and Systems (IROS)}.\hskip 1em plus 0.5em minus
  0.4em\relax IEEE, 2015, pp. 742--749.

\bibitem{krogius2019iros}
M.~Krogius, A.~Haggenmiller, and E.~Olson, ``Flexible layouts for fiducial
  tags,'' in \emph{Proceedings of the {IEEE/RSJ} International Conference on
  Intelligent Robots and Systems {(IROS)}}, October 2019.

\bibitem{lepetit2009epnp}
V.~Lepetit, F.~Moreno-Noguer, and P.~Fua, ``Epnp: Efficient perspective-n-point
  camera pose estimation,'' \emph{International Journal of Computer Vision},
  vol.~81, no.~2, pp. 155--166, 2009.

\bibitem{mobileye}
G.~Stein, O.~Mano, and A.~Shashua, ``Vision-based acc with a single camera:
  bounds on range and range rate accuracy,'' in \emph{IEEE IV2003 Intelligent
  Vehicles Symposium. Proceedings (Cat. No.03TH8683)}, 2003, pp. 120--125.

\end{thebibliography}

\end{document}